\documentclass[letterpaper, 10 pt, conference]{ieeeconf}  % Comment this line out if you need a4paper

\IEEEoverridecommandlockouts                              % This command is only needed if
\usepackage{graphics} % for pdf, bitmapped graphics files
\usepackage{amsmath} % assumes amsmath package installed
\usepackage{amssymb}  % assumes amsmath package installed
\usepackage{cite}
\usepackage{graphicx}
\usepackage{subcaption}
\usepackage{booktabs}
\usepackage{authblk}
\usepackage{url}

\title{\LARGE \bf
Reducing the Deployment-Time Inference Control Costs of \\Deep Reinforcement Learning Agents via an Asymmetric Architecture
}

\author[$\dag$]{Chin-Jui Chang}
\author[$\dag$]{Yu-Wei Chu}
\author[$\dag$]{Chao-Hsien Ting}
\author[$\dag$]{Hao-Kang Liu}
\author[$\dag$]{Zhang-Wei Hong}
\author[$\dag$]{Chun-Yi Lee}
\affil[$\dag$]{Elsa Lab, Department of Computer Science, National Tsing Hua University, Hsinchu, Taiwan}

\begin{document}

\maketitle
\thispagestyle{empty}
\pagestyle{empty}

\begin{abstract}

Deep reinforcement learning (DRL) has been demonstrated to provide promising results in several challenging decision making and control tasks. However, the required inference costs of deep neural networks (DNNs) could prevent DRL from being applied to mobile robots which cannot afford high energy-consuming computations. To enable DRL methods to be affordable in such energy-limited platforms, we propose an asymmetric architecture that reduces the overall inference costs via switching between a computationally expensive policy and an economic one. The experimental results evaluated on a number of representative benchmark suites for robotic control tasks demonstrate that our method is able to reduce the inference costs while retaining the agent's overall performance.

\end{abstract}
% \vspace{-1em}
\section{Introduction}
\label{sec::introduction}

% 強調Computational Costs開頭
% Previous works focus on a single DNN. However, some tasks cannot be solved by a single
% Motivated by the restrictions of the previous work. (1) Some tasks can be solved by a small networks
% Motivated by the problem (2) DRL computational costs (robotics?)
% The interplay 

% 

Recent works have combined reinforcement learning (RL) with the advances of deep neural networks (DNNs) to make breakthroughs in domains ranging from games~\cite{dqn, a3c, go} to robotic control~\cite{ddpg, sac, manipulate_robotic}. However, the inference phase of a DNN model is a computationally-intensive process~\cite{high_inference_cost, efficient_proc_dnn} and is one of the major concerns when applied to mobile robots, which are mostly battery-powered and have limitations on the energy budgets. % Although a number of cost-efficient techniques have been proposed \cite{deep_compression, accelerate_inference, distill_knowledge}, a cost-aware RL method is still necessary and would benefit the inference phase of the models.
Although the energy consumption of DNNs could be alleviated by reducing their sizes for energy-limited platforms, smaller DNNs are usually not able to attain same or comparable levels of performances as larger ones in complex scenarios.  On the other hand, the performances of smaller DNNs may still be acceptable in some cases.  For example, a small DNN unable to perform complex steering control is still sufficient to handle simple and straight roads.  Motivated by this observation, we propose an asymmetric architecture that selects a small DNN to act when conditions are acceptable, while employing a large one when necessary.

%While DNNs appear to be energy-demanding, reducing the size of a DNNs could help on an energy-limited platform. Though smaller networks might not attain the performance of larger DNNs, the performance can be acceptable in some cases. For example, a small network that cannot perform complex steering actions can be enough to drive straight on a road. Motivated by this observation, we propose an asymmetric architecture that selects a small network to act when acceptable and take over the small network by a large network when necessary.

% Small DNNs do not require much energy comsumption. However, they tend to deliver relatively worse performance.
% Large DNNs are able to deliver sufficient performance.
% In order to leverage the advantages of the two methods, in this paper, we propose to employ a hierarchical architecture compromising asymmetric sub-networks.

%In order to realize the above architecture, a flexible framework which is able to perform cost-aware control that leverages different amounts of efforts according to the complexities of different scenarios is necessary. 

 We implement this cost-efficient asymmetric architecture via leveraging the concept from hierarchical reinforcement learning (HRL)~\cite{hrl}, which consists of a \textit{master policy} and two \textit{sub-policies}. The master policy is designed as a lightweight DNN for decision-making, which takes in a state as its input and learns to choose a sub-policy based on the input state. The two sub-policies are separately implemented as a large DNN and a small DNN. The former is designed to deal with complicated state-action mapping, while the latter is responsible for handling simple scenarios. Therefore, when complex action control is required, the master policy uses the former. Otherwise, the latter is selected. To achieve the objective of cost-aware control, we propose a loss function design such that the inference costs of executing the two sub-policies are taken into consideration by the master policy. The master policy is required to learn to use the sub-policy with a small DNN as frequently as possible while maximizing and maintaining the agent's overall performance.
 
 Our principal contribution is an asymmetric RL architecture that reduces the deployment-time inference costs. To validate the proposed architecture, we perform a set of experiments on the representative robotic control tasks from the OpenAI Gym Benchmark Suite~\cite{openai_gym} and the DeepMind Control Suite~\cite{deepmindcontrolsuite2018}. The results show that the master policy trained by our methodology is able to alternate between the two sub-policies to save inference costs in terms of floating-point operations (FLOPs) with little performance drop. We further provide an in-depth look into the behaviors of the trained master policies, and quantitatively and qualitatively discuss why the computational costs can be reduced. Finally, we offer a set of ablation analyses to validate the design decisions of our cost-aware methodology. 
\section{Related Work}
\label{sec::related_work}

A number of knowledge distillation based methods have been proposed in the literature to reduce the inference costs of DRL agents at the deployment time~\cite{distill_knowledge, fitnets, not_need_deep, multiplier_free_dnn}.  These methods typically use a large teacher network to teach a small student network such that the latter is able to mimic the behaviors of the former.
In contrast, our asymmetric approach is based on the concept of HRL~\cite{hrl}, a framework consisting of a policy over sub-policies and a number of sub-policies for executing temporally extended actions to solve sub-tasks.
% The choice of options is flexible. Options can be either hand-crafted~\cite{hrl} or pre-trained~\cite{snn_hrl}. A number of past works~\cite{option_critic, hiro, feudal_hrl, lifelong, policy_sketches, deliberation_cost, multi_task_popart, adaptation_hrl} propose to develop options automatically, which are similar to the framework adopted by our proposed methodology.
Previous HRL works~\cite{snn_hrl, option_critic, hiro, feudal_hrl, lifelong, policy_sketches, deliberation_cost, multi_task_popart, adaptation_hrl} have been concentrating on using temporal abstraction to deal with difficult long-horizon problems. As opposed to those prior works, our proposed method focuses on employing HRL to reduce the inference costs of an RL agent. Please note that the theme and objective of the paper is  to propose a new direction of HRL to a practical problem in robot deployment scenarios, not a more general HRL strategy.

\section{Background}
\label{sec::background}

% In this section, we discuss the necessary background material of RL and HRL for understanding the primary content of this paper. We refer the interested readers to the supplementary material of this paper for additional background knowledge.

\subsection{Reinforcement Learning}
We consider the standard RL setup where an agent interacts with an environment $\mathcal{E}$ over a number of discrete timesteps, where the interaction is modeled as a Markov Decision Process (MDP). At each timestep \(t\), the agent observes a state \(s_t\) from a state space \(\mathcal{S}\), and performs an action \(a_t\) from an action space \(\mathcal{A}\) according to its policy \(\pi\), where \(\pi\) is a mapping function represented as \(\pi : \mathcal{S} \to P(\mathcal{A})\). The agent then receives the next state \(s_{t+1}\in\mathcal{S}\) and a reward signal \(r_t\) from \(\mathcal{E}\). The process continues until a termination condition is met. The objective of the agent is to maximize the expected cumulative return $\mathbb{E}[R_t] = \mathbb{E}[\Sigma_{k=0}^{\infty} \gamma^{k} r_{t+k}]$ from \(\mathcal{E}\) for each timestep \(t\), where $\gamma\in (0, 1]$ is a discount factor.  % Further details of RL can be referred in the supplementary material.

\subsection{Hierarchical Reinforcement Learning}
HRL introduces the concept of `options' into the RL framework, where options are temporally extended actions. 
\cite{hrl} shows that an MDP combined with options becomes a Semi-Markov Decision Process (SMDP). 
Assume that there exists a set of options $\Omega$. HRL allows a `policy over options' $\pi_{\Omega}$ to determine an option for execution for a certain amount of time. Each option \(\omega\in\Omega\) consists of three components \((\mathcal{I}_\omega, \pi_\omega, \mathcal{\beta}_\omega)\), in which \(\mathcal{I}_\omega\subseteq \mathcal{S}\) is an initial set according to $\pi_{\Omega}$, \(\pi_\omega\) is a policy following option \(\omega\), and \(\mathcal{\beta}_\omega:\mathcal{S}\to[0, 1]\) is a termination function. When an agent enters a state \(s\in \mathcal{I}_\omega\), option \(\omega\) is adopted, and policy \(\pi_\omega\) is followed until a state \(s_k\) where \(\beta_\omega(s_k)\to1\). In episodic tasks, termination of an episode also terminates the current option. Our architecture is a special case of SMDP. Section~\ref{sec::methodology} introduces our update rules for \(\pi_\Omega\) and \(\pi_\omega\). In this paper, we refer to a `policy over options' as a \textit{master policy}, and an `option' as a \textit{sub-policy}.

\section{Methodology}
\label{sec::methodology}

% In this section, we first introduce the problem formulation. Then, we describe our framework and the training method. % The pseudo-code is provided in the supplementary material.
% \vspace{-0.5em}

\begin{figure}[t]
    \centering
    \includegraphics[width=\linewidth]{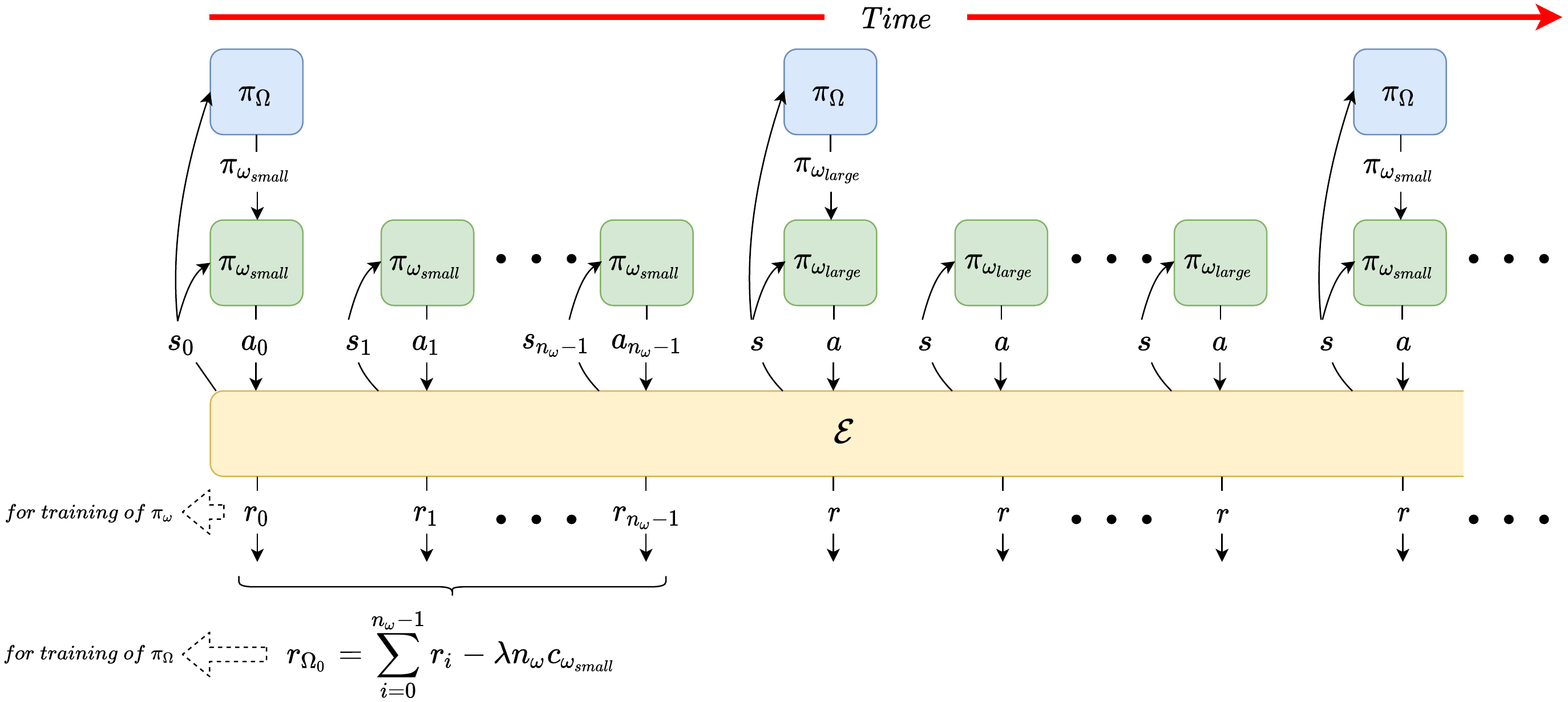}
    \caption{An illustration of the workflow of our framework. The master policy $\pi_\Omega$ chooses a sub-policy $\pi_\omega \in \{\pi_{\omega_{small}}, \pi_{\omega_{large}}\}$, and uses it to interact with the environment $\mathcal{E}$ for $n_\omega$ timesteps. After this, $\pi_\Omega$ chooses another $\pi_\omega$, and the process repeats until the end of the episode. $\pi_\Omega$ and $\pi_\omega$ use different experience transitions to update their policies and have different replay buffers, while $\pi_{\omega_{small}}$ and $\pi_{\omega_{large}}$ share the same replay buffer. We train $\pi_{\omega_{small}}$ and $\pi_{\omega_{large}}$ with experience transitions $(s_t, a_t, r_t, s_{t+1}) \text{ for } t=0, 1, ..., etc$, and train $\pi_\Omega$ with transitions $(s_t, \omega_t, r_{\Omega_t}, s_{t+n_\omega}) \text{ for } t=0$, $\,n_\omega, ..., etc$. The reward $r_{\Omega_t}=\Sigma_{i=t}^{t+n_\omega-1} r_i\allowbreak-\lambda n_\omega c_{\omega_t}$ of $\pi_\Omega$ is the sum of rewards $r_i$ collected by $\pi_\omega$, which is penalized by the scaled cost $\lambda n_\omega c_{\omega_t}$ of $\pi_\omega$. Note that  $\lambda$ is a scaling parameter. }
    \label{fig:work_flow}
\end{figure}

\subsection{Problem Formulation}
\label{subsec::problem_formulation}

The main objective of this research is to develop a cost-aware strategy such that an agent trained by our methodology is able to deliver satisfying performance while reducing its overall inference costs. 
We formulate the problem as an SMDP, with an aim to train the master policy in the proposed framework to use the smaller sub-policy when the condition is appropriate to be handled by it, and employ the larger sub-policy when the agent requires complex control of its actions.  The agent is expected to use the smaller sub-policy as often as possible to reduce its computational costs.
% We formulate the problem as an SMDP, and hypothesize that an episode of a control task can be decomposed into $K$ segments, where each segment $Seg^k, \forall k\in K$ can be categorized to either a simple segment $Seg_{simple}$ or a difficult segment $Seg_{difficult}$ according to their required control complexities, and is represented as $Seg^k\in \{{Seg}_{simple}, {Seg}_{difficult}\}$. The agent is expected to use a simpler sub-policy with a smaller DNN when handling a segment $Seg^i\in \{{Seg}_{simple}\}, \forall i\in K$, while choosing a complex sub-policy with a larger DNN when dealing with a segment $Seg^j\in \{{Seg}_{difficult}\}, \forall j\in K$, where $i\neq j$. The decision of the sub-policies to be used by the agent for different segments is determined by another policy.  
% We formulate the problem as an SMDP, and hypothesize that an episode of a control task can be decomposed into $K$ segments, where each segment $Seg^k, \forall k\in K$ can be categorized to either a simple segment $Seg_{simple}$ or a difficult segment $Seg_{difficult}$ according to their required control complexities, and is represented as $Seg^k\in \{{Seg}_{simple}, {Seg}_{difficult}\}$. The agent is expected to use a simpler sub-policy with a smaller DNN when handling a segment $Seg^i\in \{{Seg}_{simple}\}, \forall i\in K$, while choosing a complex sub-policy with a larger DNN when dealing with a segment $Seg^j\in \{{Seg}_{difficult}\}, \forall j\in K$, where $i\neq j$. The decision of the sub-policies to be used by the agent for different segments is determined by another policy. 
In order to incorporate the consideration of inference costs into our cost-aware strategy, we further assume that each sub-policy is cost-bounded. The cost of a sub-policy 
% is proportional to the number of FLOPs for performing inference and 
is denoted as $c_{\omega}$, where $\omega$ represents the sub-policy used by the agent. The reward function is designed such that the agent is encouraged to select the lightweight sub-policy as frequently as possible to avoid being penalized.

\subsection{Overview of the Cost-Aware Framework}
% \vspace{-0.5em}
In order to address the problem formulated above, we employ an HRL framework consisting of a master policy $\pi_\Omega$ and two sub-policies $\pi_\omega$ of different DNN sizes, where $\omega \in \{\omega_{small}, \omega_{large}\}$ and the DNN size of $\omega_{large}$ is larger than that of $\omega_{small}$. We assume that both the sub-policies $\pi_\omega$ can be completed in a single timestep. At the beginning of a task, $\pi_\Omega$ first takes in the current state $s\in \mathcal{S}$ from $\mathcal{E}$ to determine which $\pi_\omega$ to use. The selected $\pi_\omega$ is then used to interact with $\mathcal{E}$ for $n_\omega$ timesteps, i.e., $\mathcal{\beta}_\omega\to1$ once the selected sub-policy $\omega$ is used for $n_\omega$ timesteps. The value of $n_\omega$ is set to be a constant for the two sub-policies, i.e., \(n_{\omega_{large}}=n_{\omega_{small}}\). The process repeats until the end of the episode. The workflow of the proposed cost-aware hierarchical framework is illustrated in Fig.~\ref{fig:work_flow}. Please note that even though the overall system is formulated as an SMDP, the formulation for $\pi_{\Omega}$ is still a standard MDP problem of selecting between a set of two temporally extended actions (i.e., using either \(\pi_{\omega_{small}}\) or \(\pi_{\omega_{large}}\)), as described in Section 3 of \cite{hrl}. Therefore, at timestep $t$, the goal of $\pi_\Omega$ becomes maximizing \(R_{\Omega_t} = \Sigma_{i=0}^{\infty} \gamma^{i} r_{\Omega_{t+i\cdot n_\omega}}\), where \(r_{\Omega_t}=\Sigma_{j=t}^{t+n_\omega-1} r_j\) is the cumulative rewards during the execution of \(\pi_\omega\). On the other hand, the update rule of \(\pi_\omega\) is the same as the intra-option policy gradient described in \cite{option_critic}. To deal with the data imbalance issue of the two sub-policies during the training phase as well as improving data efficiency, our cost-aware framework uses an off-policy RL algorithm for \(\pi_\omega\) so as to allow \(\pi_{\omega_{small}}\) and \(\pi_{\omega_{large}}\) to share the common experience replay buffer. % In order to validate this design decision, we provide additional experimental results with and without using a shared experience replay buffer across \(\pi_\omega\) in the supplementary material.

\subsection{Cost-Aware Training}
% \vspace{-0.5em}
% In this subsection, we introduce how we train \(\pi_\Omega\) and \(\pi_\omega\). For \(\pi_\Omega\), the problem is formulated as an SMDP. However, for options with identical transition time, the problem becomes a standard MDP problem (section 2 in \cite{rl_continous_time}). We can view \(\pi_\Omega\) as choosing from a different set of actions, i.e., using either \(\pi_{\omega_{small}}\) or \(\pi_{\omega_{large}}\). Therefore the goal \(\pi_\Omega\) wants to maximize becomes \(R_{\Omega_t} = \Sigma_{k=0}^{\infty} \gamma^{k} r_{\Omega_{t+k\cdot n_\omega}}\) where \(r_{\Omega_t}=\Sigma_{i=t}^{t+n_\omega-1} r_i\) is the accumulative reward during the execution of \(\pi_\omega\). For \(\pi_\omega\), the update rule is the same as the standard RL. However, in order to solve the unbalanced training problem and improve data efficiency, we choose to use an off-policy RL algorithm for \(\pi_\omega\), so \(\pi_\omega\) can share their experience replay buffer. We also compare our results to models without sharing replay buffer across \(\pi_\omega\) in section \ref{sec:ablation}.

We next describe the training methodology. In case that no regularization is applied, $\pi_\Omega$ tends to choose $\pi_{\omega_{large}}$ due to its inherent advantages of being able to obtain more rewards on its own. As a result, we penalize $\pi_\Omega$ with \(c_\omega\) to encourage it to choose $\pi_{\omega_{small}}$ with a lower \(c_\omega\). The reward for $\pi_\Omega$ at $t$ is thus modified to \(r_t - \lambda c_\omega\), where \(\lambda\) is a cost coefficient for scaling. The higher the value of \(\lambda\) is, the more likely \(\pi_\Omega\) will choose $\pi_{\omega_{small}}$. The experience transitions used to update $\pi_\Omega$ are therefore expressed as \((s_t, \omega_t, r_{\Omega_t}, s_{t+n_\omega}) \text{ for } t=0\), \(n_\omega, 2n_\omega, ..., etc\), where \(r_{\Omega_t}=\Sigma_{i=t}^{t+n_\omega-1} r_i-\lambda n_\omega c_{\omega_t}\). % The pseudo-code of the training methodology as well as the hyparameter searching procedure of $\lambda$ are provided in the supplementary material. 

\section{Experimental Results}
\label{sec::experimental_results}

\subsection{Experimental Setup}
\label{sec:experimental_setup}

% We first present the environments used in our experiments in Section~\ref{subsubsed::environments}, followed by a brief description of the hyperparameters in Section~\ref{subsubsed::hyperparameters}. Next, we describe our baselines in Section~\ref{subsubsed::baselines}. 

% To practically quantify computational costs, in our experiments, we adopt the inference FLOPs of \(\pi_\omega\) as \(c_\omega\). 

%The curves presented in the experiments are generated based on five random seeds and drawn with 95\% confidence interval as the shaded areas.

\subsubsection{Environments}
\label{subsubsed::environments}
We verify the proposed methodology in simple classic control tasks from the OpenAI Gym Benchmark Suite~\cite{openai_gym}, and a number of challenging continuous control tasks from both the OpenAI Gym Benchmark Suite and the DeepMind Control Suite~\cite{deepmindcontrolsuite2018} simulated by the MuJoCo~\cite{mujoco} physics engine. The challenging tasks include four continuous control tasks from the OpenAI Gym Benchmark Suite, and two tasks from the DeepMind Control Suite. % We present several representative results in this section, and summarize the evaluation results as well as the analyses of the remaining control tasks in our supplementary material.
% \textcolor{red}{We have additional experimental results performed on a real robot in the supplementary material.}

%We test our method on simpler classic control tasks such as \textit{MountainCar} and challenging continuous control tasks from the OpenAI gym benchmark suite \cite{openai_gym}, including 10 simulated robotics tasks which use the MuJoCo \cite{mujoco} physics engine. 

\subsubsection{Hyperparameters}
\label{subsubsed::hyperparameters}
% Seeds

\begin{table}[t]
% \begin{table}[t]
\centering
\renewcommand{\arraystretch}{1.1}
\caption{
The detailed settings of the hyperparameters adopted by the master policy $\pi_\Omega$ and the sub-policies $\pi_\omega$ of our methodology.
}
\label{tab:ours_hyperparam}
\footnotesize
\resizebox{.8\columnwidth}{!}{
\begin{tabular}{lc|lc}
\toprule \toprule
\multicolumn{1}{c}{Hyperparameter}                                      & Value & \multicolumn{1}{c}{Hyperparameter}                                      & Value  \\ \toprule \toprule
\multicolumn{2}{c|}{\textbf{Master Policy $\pi_\Omega$}} & \multicolumn{2}{c}{\textbf{Sub-Policy $\pi_\omega$}}
            \\ \midrule
            RL algorithm & DQN & RL algorithm & SAC \\
            Learning rate & $1\mathrm{e}{-3}$ & Entropy coefficient $\alpha$ & Auto \\
    		Discount factor ($\gamma$) & $0.99$ & Learning rate of agent & $3\mathrm{e}{-4}$ \\
    		Replay buffer size & 50K & Discount factor ($\gamma$) & $0.99$ \\
    		Exploration fraction & 10\% & Replay buffer size & 1M \\ 
    		Update batch size & 32 & Update batch size & 256 \\
    		Double Q & True & Train frequency & 1 \\ 
    		Train frequency & 1 & Target soft update coefficient $\tau$ & 0.005 \\
    		Target network update interval & 500 & Target network update interval & 1 \\
            Optimization for the RL agent & Adam & Optimization for the RL agent & Adam \\
            Training timesteps & 2.5M & Training timesteps & 2.5M \\
            Nonlinearity & Tanh & Nonlinearity & Tanh \\
%             \toprule
% \multicolumn{2}{c}{\textbf{Sub-Policy $\pi_\omega$}}
%             \\ \midrule
%             RL algorithm & SAC \\
%             Entropy coefficient $\alpha$ & Auto \\
%     		Learning rate of agent & $3\mathrm{e}{-4}$ \\
%     		Discount factor ($\gamma$) & $0.99$ \\
%     		Replay buffer size & 1M \\
%     		Update batch size & 256 \\
%     		Train frequency & 1 \\
%     		Target soft update coefficient $\tau$ & 0.005 \\ 
%     		Target network update interval & 1 \\
%             Optimization for the RL agent & Adam \\
%             Training timesteps & 2.5M \\
%             Nonlinearity & Tanh \\
\bottomrule \bottomrule
\end{tabular}
}
% \end{table}
\end{table}

\begin{table}[t]
  \caption{Number of neurons $n_{units}$ per layer for $\pi_{\omega_{small}}$ \& $\pi_{\omega_{large}}$, $c_{\omega_{small}}$, $c_{\omega_{large}}$, and $\lambda$ for each robotic control tasks.}
  \label{tab:n_units}
  \centering
  \renewcommand{\arraystretch}{1.1}
  \small
  \resizebox{\columnwidth}{!}{
      \begin{tabular}{c|cc|ccc}
        \toprule
        Environment & $n_{units}$ for $\pi_{\omega_{small}}$ & $n_{units}$ for $\pi_{\omega_{large}}$ & $c_{\omega_{small}}$ & $c_{\omega_{large}}$  & $\lambda$\\
        \midrule
        \textit{MountainCarContinuous-v0} & $8$ & $64$ & $1.0$ & $44.7$ & $1\mathrm{e}{-4}$\\
        \textit{Swimmer-v3} & $8$ & $256$ & $1.0$ & $428.4$ & $1\mathrm{e}{-4}$\\ 
        \textit{Ant-v3} & $64$ & $256$ & $1.0$ & $8.0$ & $1\mathrm{e}{-1}$ \\
        \textit{FetchPickAndPlace-v1} & $32$ & $128$ & $1.0$ & $9.4$ & $2\mathrm{e}{-4}$\\
        \textit{walker-stand} & $8$ & $64$ & $1.0$ & $18.1$ & $1\mathrm{e}{-2}$\\
        \textit{finger-spin} & $8$ & $64$ & $1.0$ & $29.1$ & $1\mathrm{e}{-2}$\\ 
        \bottomrule
      \end{tabular}
  }
  %\vspace{-1em}
\end{table}

In our experiments, the master policy $\pi_\Omega$ is implemented as a Deep Q-Network (DQN)~\cite{dqn} agent to discretely choose between the two sub-policies. On the other hand, the sub-policies $\pi_\omega$ are implemented as Soft Actor-Critic (SAC)~\cite{sac} agents for performing the continuous control tasks described above. The hyperparameters used for training are shown in Table~\ref{tab:ours_hyperparam}. Both $\pi_\Omega$ and $\pi_\omega$ are implemented as multilayer perceptrons (MLPs) with two hidden layers. We set the number of units $n_{units}$ per layer for $\pi_\Omega$ to 32 for all tasks, and determine $n_{units}$ for $\pi_\omega$ as follows. We first train a model with $n_{units}$ set to 512 as the criterion model, and then find the minimum $n_{units}$ for the model which can achieve $90\%$ of the performance of the criterion model. We use this as $n_{units}$ for $\pi_{\omega_{large}}$. And then we find $n_{units}$ for $\pi_{\omega_{small}}$, such that its value is less than or equal to $1/4$ of $n_{units}$ for $\pi_{\omega_{large}}$ and the performance of $\pi_{\omega_{small}}$ is around or below $1/3$ of the score achieved by the criterion model. 

\begin{figure}[t]
  \centering
  \includegraphics[width=.95\linewidth]{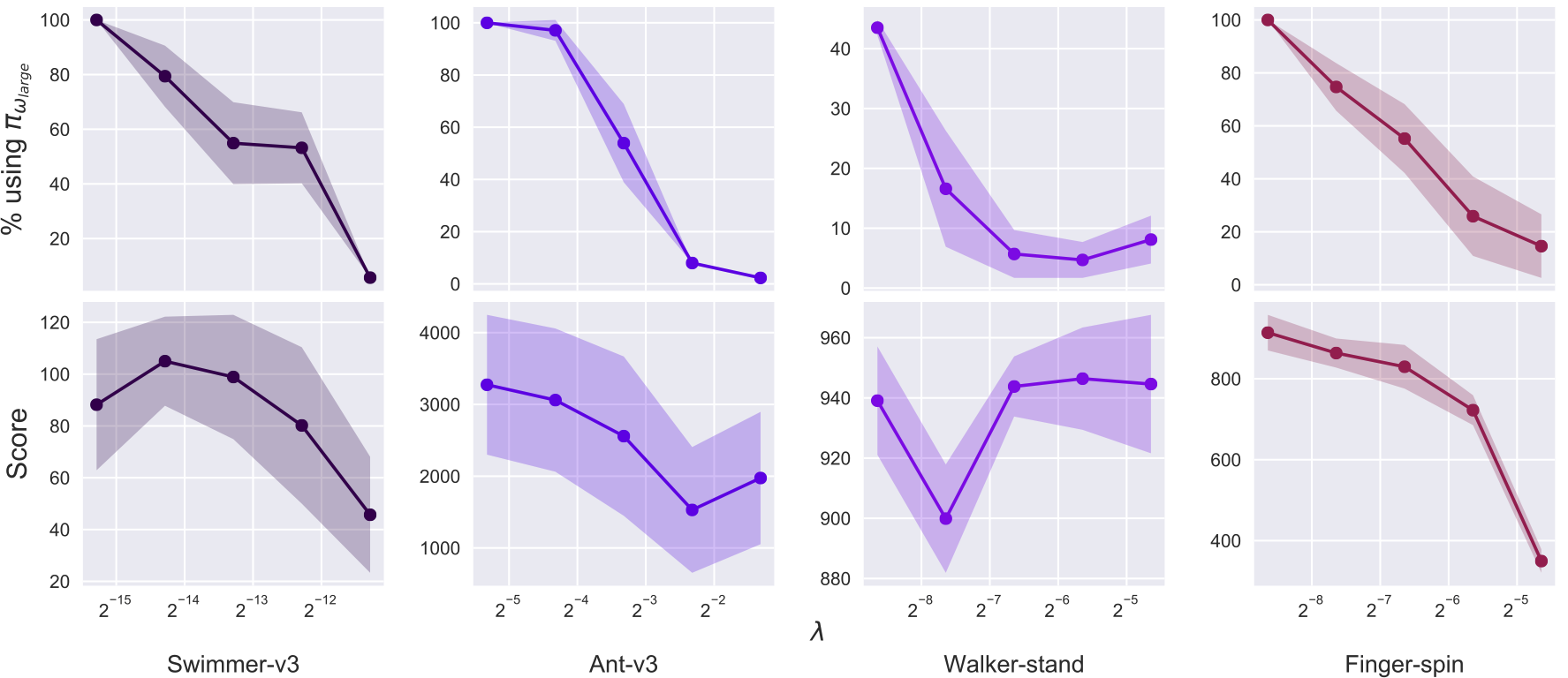}
  \caption{Performance of the models trained with different $\lambda$. The scores are averaged from 5 different random seeds. Each model trained with different random seed is evaluated over 200 episodes.}
  \label{fig:diff_lambda}
\end{figure}

For the cost term, we adopt the inference FLOPs of \(\pi_\omega\) as \(c_\omega\), since the FLOPs executed by \(\pi_\omega\) and its energy consumption are correlated. We use the number of FLOPs of $\pi_{\omega_{small}}$ and $\pi_{\omega_{large}}$ divided by the number of FLOPs of $\pi_{\omega_{small}}$ as their policy costs $c_{\omega_{small}}$ and $c_{\omega_{large}}$, respectively, such that $c_{\omega_{small}}$ is equal to one. With regard to $\lambda$, from Fig.~\ref{fig:diff_lambda}, we observe that $\lambda$ and the ratio of choosing $\pi_{\omega_{large}}$ is negatively correlated. Even though the performances decline along with the reduced usage rate of $\pi_{\omega_{large}}$, there is often a range of $\lambda$ which leads to lower usage rate of $\pi_{\omega_{large}}$ and yet comparable performance to the model of high $\pi_{\omega_{large}}$ usage rate. We perform a hyperparameter search to find an appropriate $\lambda$, such that both $\pi_{\omega_{small}}$ and $\pi_{\omega_{large}}$ are used alternately within an episode, while allowing the agent to obtain high scores. The hyperparameters used in the cost term are listed in Table~\ref{tab:n_units}.

\begin{figure}[t]
  \centering
  \includegraphics[width=.95\linewidth]{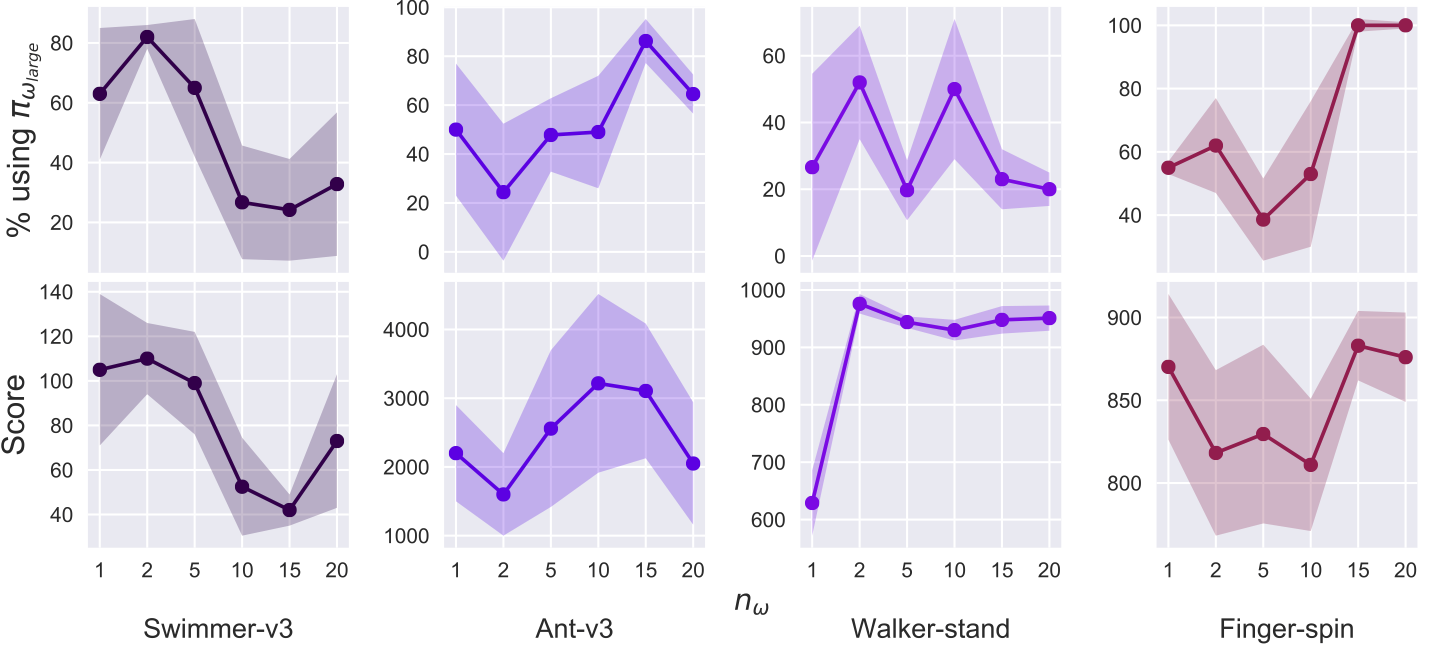}
  \caption{Performance of models trained with different $n_\omega$. The scores are averaged from 5 different random seeds. Each model trained with different random seed is evaluated over 200 episodes.}
  \label{fig:diff_nomega}
\end{figure}
We also do hyperparameter searches to find $n_\omega$. It can be observed from Fig.~\ref{fig:diff_nomega} that there is no obvious correlation between $n_\omega$ and the performance. \textit{Swimmer-v3} performs well with smaller values of $n_\omega$, while \textit{walker-stand} performs well with larger values of $n_\omega$. On the other hand, \textit{Ant-v3} performs well with $n_\omega$ equal to around $10$. Therefore, the choice of $n_\omega$ is relatively non-straightforward. We select the value of $n_\omega$ on account of two considerations: (1) $n_\omega$ should not be too small, or it will lead to increased master policy costs due to more frequent inferences of the master policy to decide which sub-policy to be used next; (2) $n_\omega$ should not be too large, otherwise the model will not be able to perform flexible switching between sub-policies. As a result, we set $n_\omega$ to five for all of the experiments considered in this work as a compromise. However, please note that an adaptive scheme of the step size $n_\omega$ may potentially further enhance the overall performance, and is left as a future research direction.

For \textit{FetchPickAndPlace-v1}, we train the model with hindsight experience replay (HER)~\cite{her} to improve the sample efficiency. For most of the results, the default training and evaluation lengths are set to 2.5M timesteps and 200 episodes, respectively. The agents are implemented based on the source codes from Stable Baselines~\cite{stable-baselines} as well as RL Baselines Zoo~\cite{rl-zoo}, and are trained using five different random seeds. 
%  The hyperparameter setups for training our agents and the baselines, as well as the derivation procedure of $\lambda$, are provided in the supplementary material.  

\subsubsection{Baselines}
\label{subsubsed::baselines}

The baselines considered include two categories: (1) a typical RL method, and (2) distillation methods.

\textbf{Typical RL method.} 
To study the performance drop and the cost reduction compared with standard RL methods, we train two policies of different sizes (i.e., numbers of DNN parameters), where the small one and the large one are denoted as $\pi_{S-only}$ and $\pi_{L-only}$ respectively. 
The sizes of $\pi_{S-only}$ and $\pi_{L-only}$ correspond to $\pi_{\omega_{small}}$ and $\pi_{\omega_{large}}$ used in our method.
Both $\pi_{S-only}$ and $\pi_{L-only}$ are trained independently from scratch as typical RL methods without the use of $\pi_{\Omega}$.

\textbf{Distillation methods.} 
In order to study the effectiveness on cost reduction, we compare our methodology with a commonly used method in RL: policy distillation. Two policy distillation approaches are considered in our experiments: Behavior Cloning (BC)~\cite{bc_limitation} and Generative Adversarial Imitation Learning (GAIL)~\cite{gail}. For these baselines, a costly policy (i.e., the large policy) serves as the teacher model that distills its policy to an economic policy (i.e., the small policy). In our experiments, the teacher network is set to $\pi_{{L-only}}$, while the configurations of the student networks are described in Section~\ref{sec:baseline}. Please note that these baselines require more training data than the typical RL method baselines and our methodology, since they need data samples from expert (i.e., $\pi_{{L-only}}$) trajectories for training their student networks.

\begin{figure}[t]
\centering
\hspace{0.2em}
\begin{subfigure}{.9\linewidth}
  \centering
  \includegraphics[width=\linewidth]{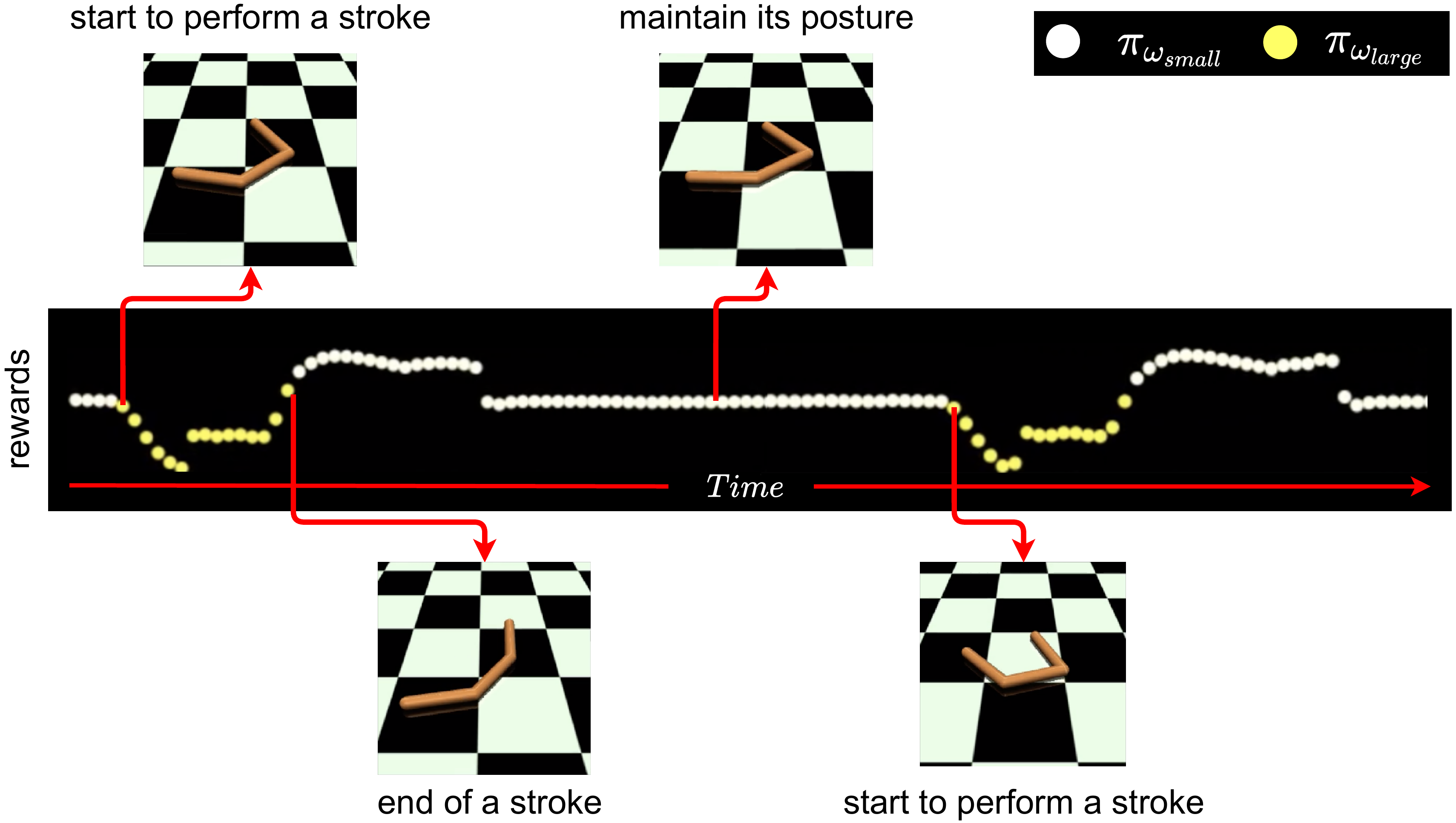}
  % \caption{Swimmer-v3}
\end{subfigure}%
\newline
\begin{subfigure}{.9\linewidth}
    \centering
      \includegraphics[width=.6\linewidth]{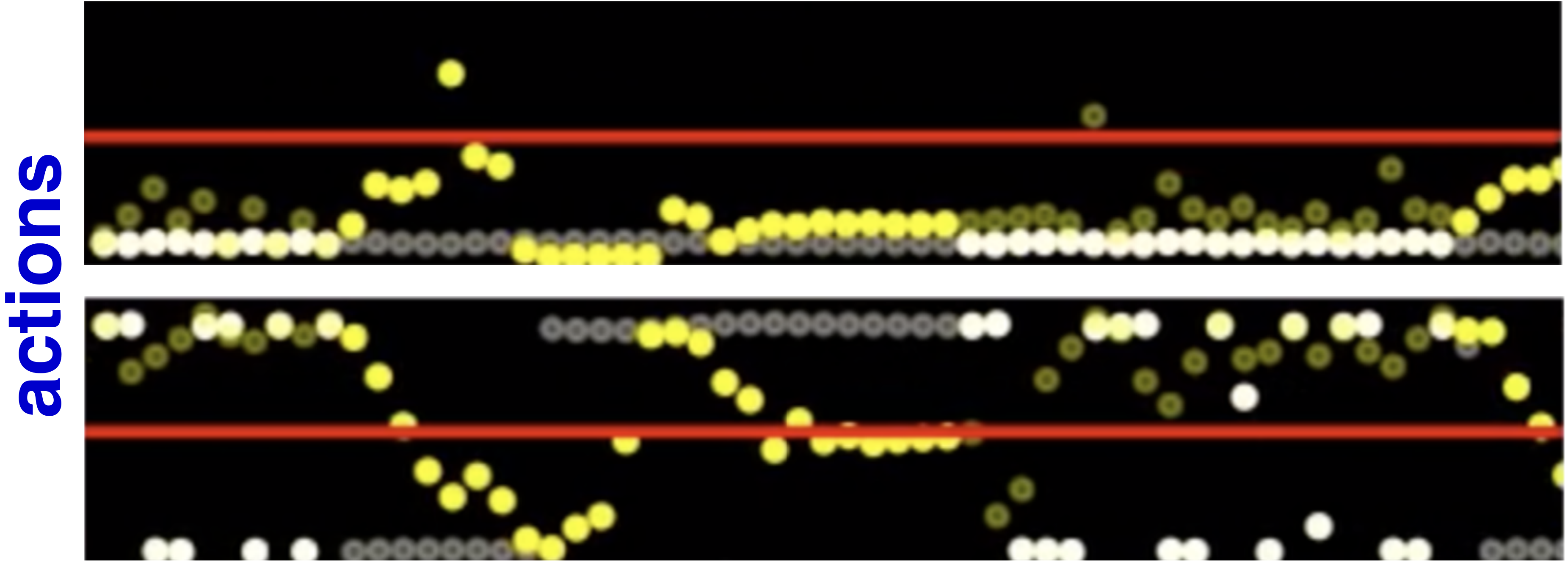}
\end{subfigure}
\caption{
%Timeline plot to demonstrate the switch between small and large policy. 
A timeline for illustrating the sub-policies used for different circumstances in \textit{Swimmer-v3},
where the interleavedly plotted white and yellow dots along the timeline correspond to the sub-policies $\pi_{\omega_{small}}$ and $\pi_{\omega_{large}}$, respectively. The master policy $\pi_{\Omega}$ selects $\pi_{\omega_{large}}$ when performing strokes, and employs $\pi_{\omega_{small}}$ to maintain or slightly adjust the posture of the swimmer between two strokes for drifting. The image at the bottom shows the actions conducted by the model. The transparent dots are actions decided by the not selected sub-policy. The opaque and the transparent dots reveal that actions conducted by $\pi_{\omega_{large}}$ is more complicated than $\pi_{\omega_{small}}$.\\
\vspace{-1em}
%The master policy chooses $\pi_{\omega_{small}}$ for most of the time to save the computational costs, and selects $\pi_{\omega_{large}}$ to deliver comparable performance to to the $\pi_{\omega{large-only}}$ baseline.\\
}
\label{fig:timeline_swimmer}
% \vspace{-2em}
\end{figure}

\begin{figure}[t]
\begin{subfigure}{.6\linewidth}
  \includegraphics[width=\linewidth]{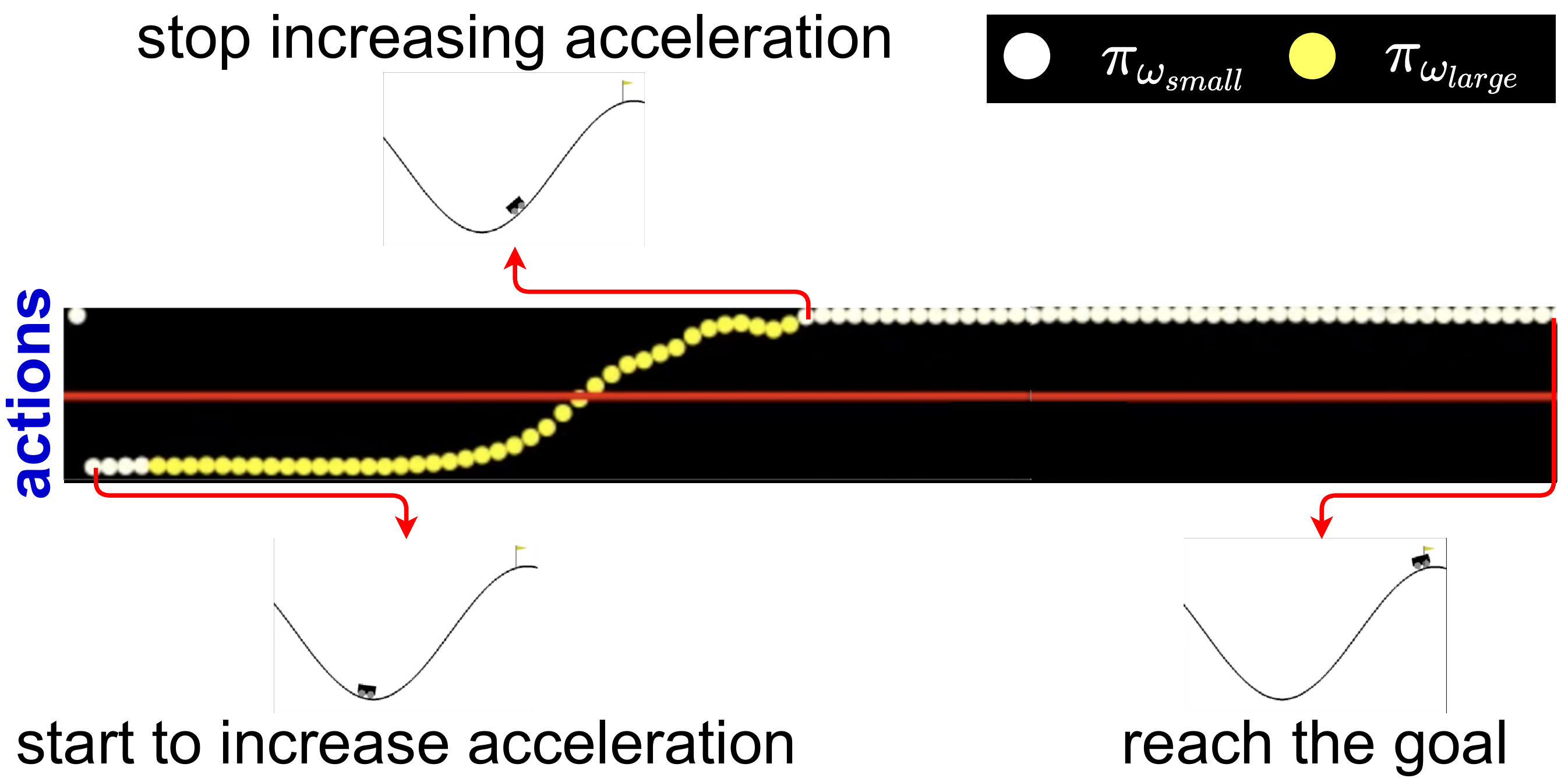}
  \caption{\textit{MountainCarContinuous-v0}}
  \label{fig:timeline_mcar}
\end{subfigure}%
% \vspace{1.5em}
%\newline
\begin{subfigure}{.4\linewidth}
  \centering
  \includegraphics[width=\linewidth]{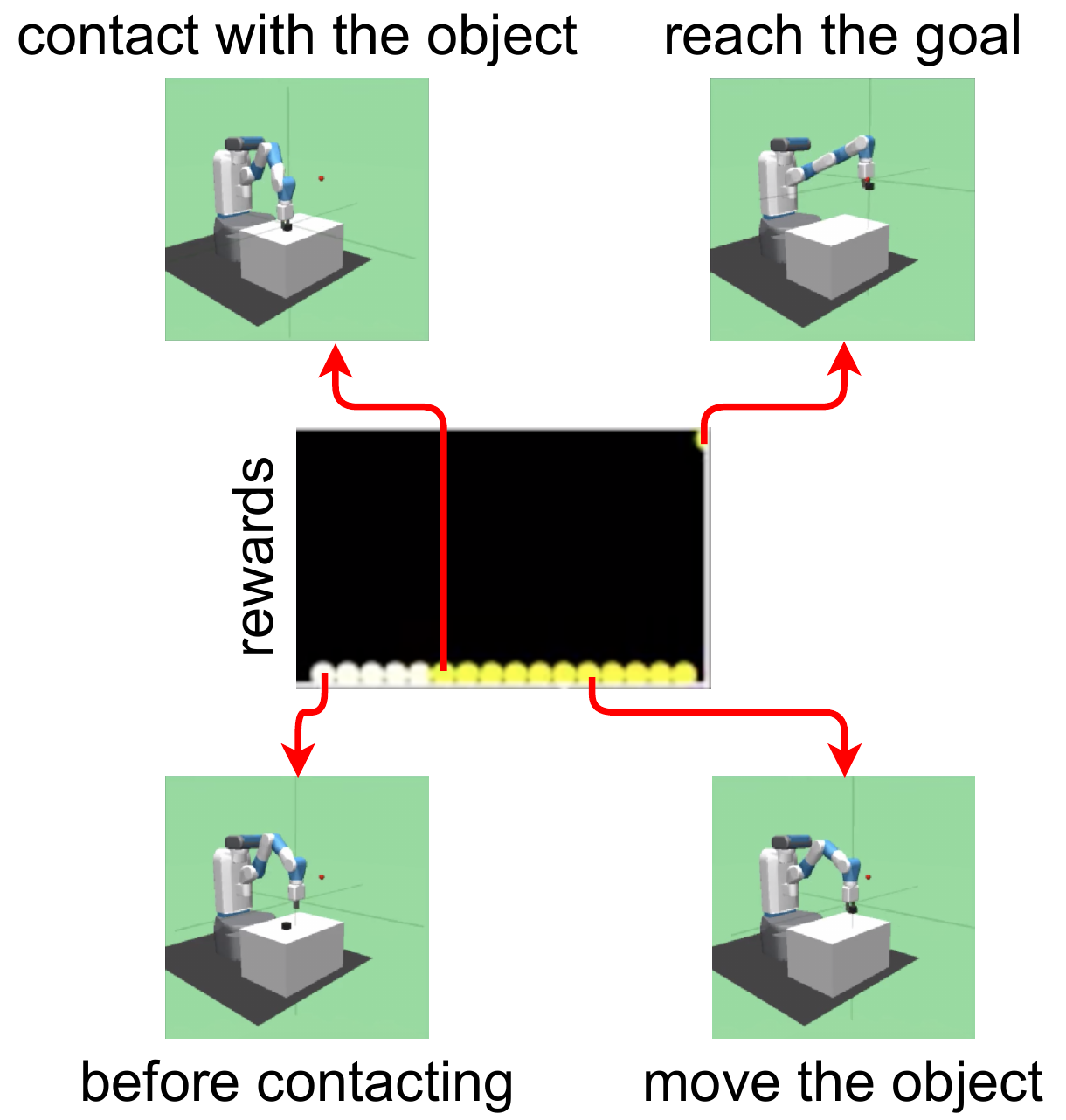}
  \caption{\textit{FetchPickAndPlace-v1}}
  \label{fig:timeline_fpap}
\end{subfigure}%
\newline
%\centering
% \begin{subfigure}{.9\textwidth}
%   \centering
%   \includegraphics[width=\linewidth]{figures/timeline_plots/timeline-cartpole-swingup.pdf}
%   \caption{\textit{Cartpole-swingup}}
%   \label{fig:timeline_cartpole_swingup}
% \end{subfigure}%
% \vspace{1.5em}
\newline
\centering
\begin{subfigure}{\linewidth}
  \centering
  \includegraphics[width=\linewidth]{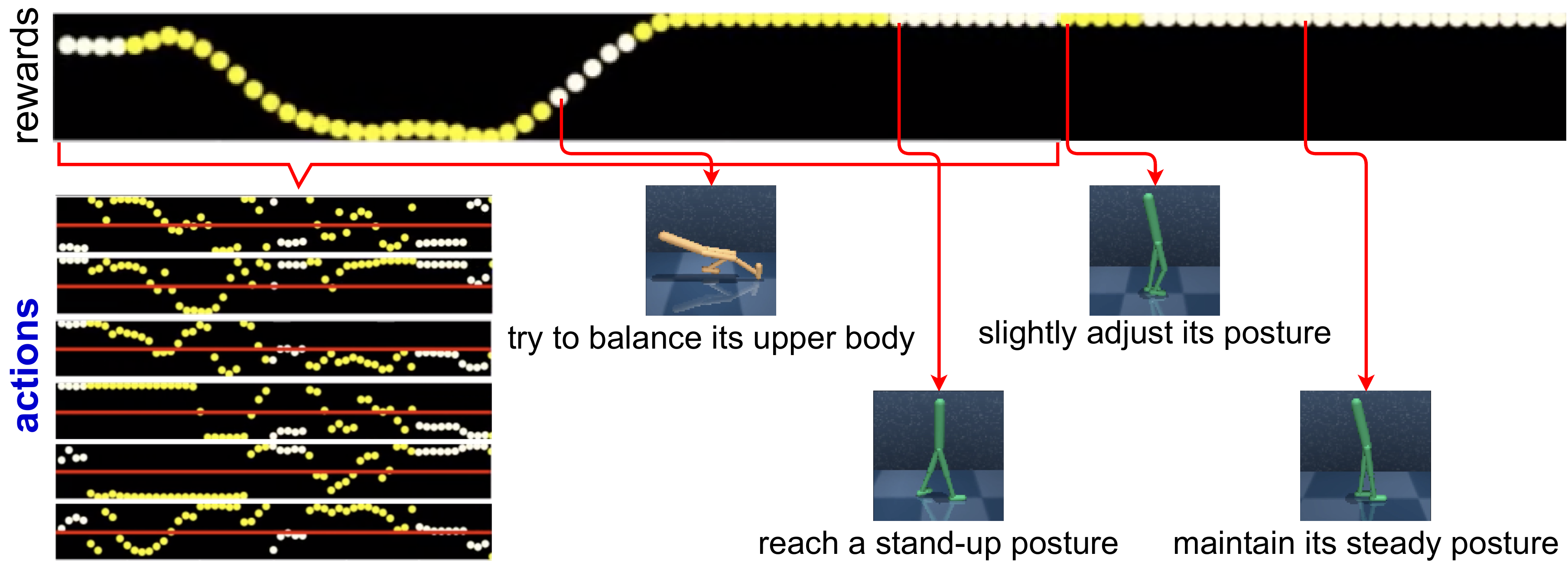}
  \caption{\textit{Walker-stand}}
  \label{fig:timeline_walker_stand}
\end{subfigure}%
\vspace{.5em}
\newline
\centering
\begin{subfigure}{\linewidth}
  \centering
  \includegraphics[width=\linewidth]{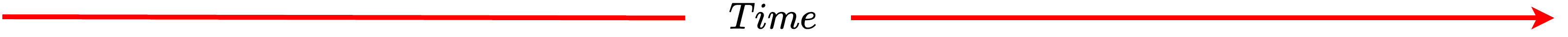}
\end{subfigure}%

\caption{\subref{fig:timeline_mcar} The mountain car uses $\pi_{\omega_{large}}$ to adjust its acceleration from a negative value to a positive value, while using $\pi_{\omega_{small}}$ to maintain its acceleration. \subref{fig:timeline_fpap} The robotic arm first approaches the object using $\pi_{\omega_{small}}$, and then employs $\pi_{\omega_{large}}$ to move the object to the target location. \subref{fig:timeline_walker_stand} The walker first utilizes $\pi_{\omega_{large}}$ and $\pi_{\omega_{small}}$ alternately to stand up. After reaching an upright posture, the walker leverages $\pi_{\omega_{small}}$ to maintain it afterwards.
}
% (\subref{fig:timeline_cartpole_swingup}) The pole uses $\pi_{\omega_{large}}$ to increase its rotation speed, while using $\pi_{\omega_{small}}$ to decelerate during the process. After the pole reaches a steady upright posture, $\pi_{\omega_{small}}$ is used to maintain it afterwards.

%(the values of white dots in the action pattern on the bottom-left change much slower than the yellow dots)
% A timeline plot for illustrating the sub-policies used for different task segments in \textit{Swimmer-v3}, where the interleavedly plotted white and yellow dots along the timeline correspond to the sub-policies $\pi_{\omega_{small}}$ and $\pi_{\omega_{large}}$, respectively. The master policy $\pi_{\Omega}$ selects $\pi_{\omega_{large}}$ when performing strokes, and uses $\pi_{\omega_{small}}$ to slightly adjust the posture of the swimmer between two strokes. Additional examples and analyses for the other control tasks are provided in the supplementary material.
\label{fig:timeline}
\end{figure}

\begin{figure*}[t]
\centering
\begin{subfigure}{.25\textwidth}
  \centering
  \includegraphics[width=\linewidth]{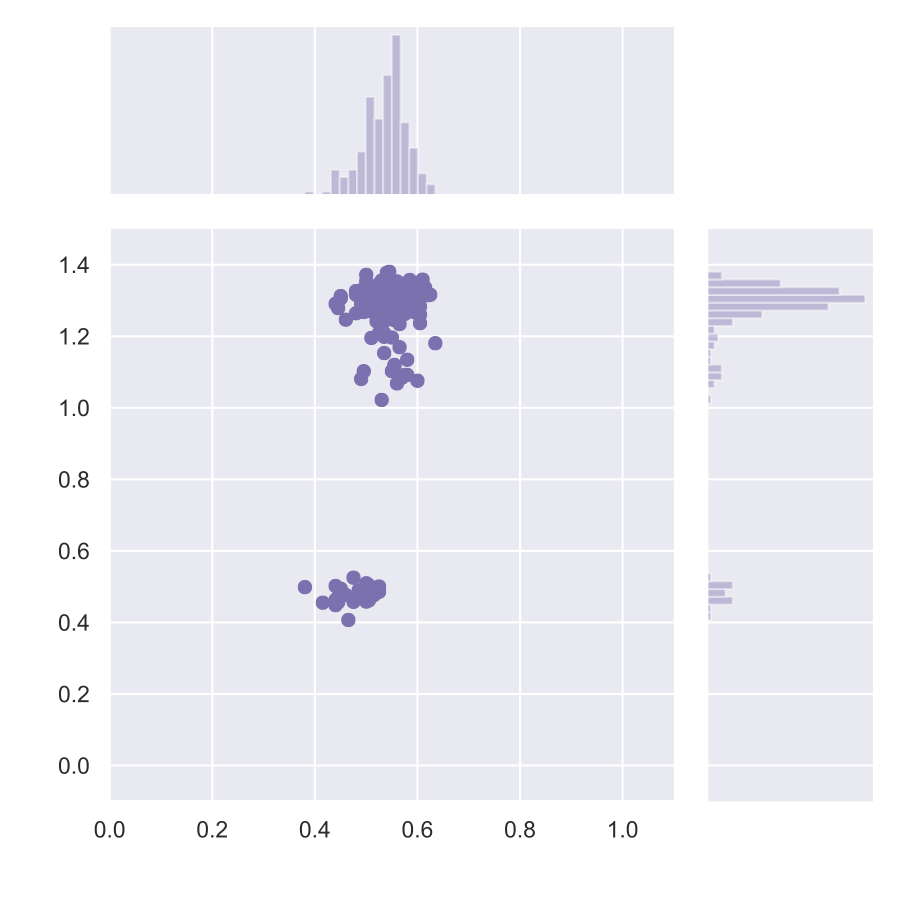}
  \caption{\textit{Swimmer-v3}}
\end{subfigure}%
\begin{subfigure}{.25\textwidth}
  \centering
  \includegraphics[width=\linewidth]{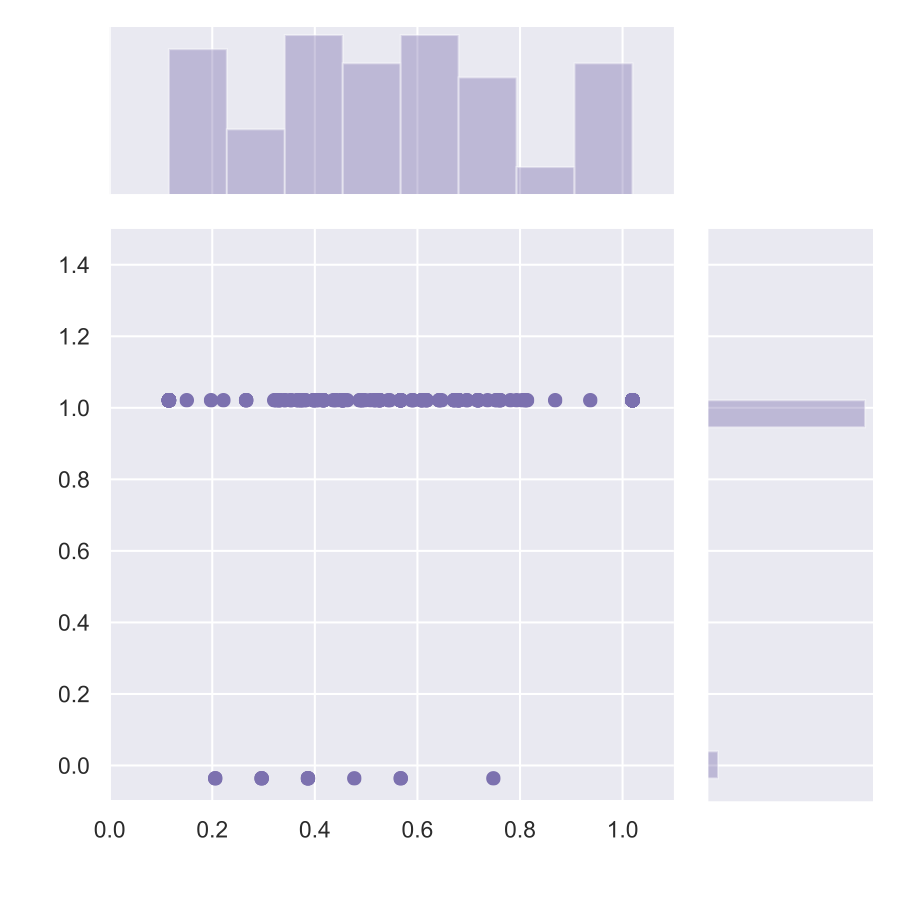}
  \caption{\textit{FetchPickAndPlace-v1}}
\end{subfigure}%
\begin{subfigure}{.25\textwidth}
  \centering
  \includegraphics[width=\linewidth]{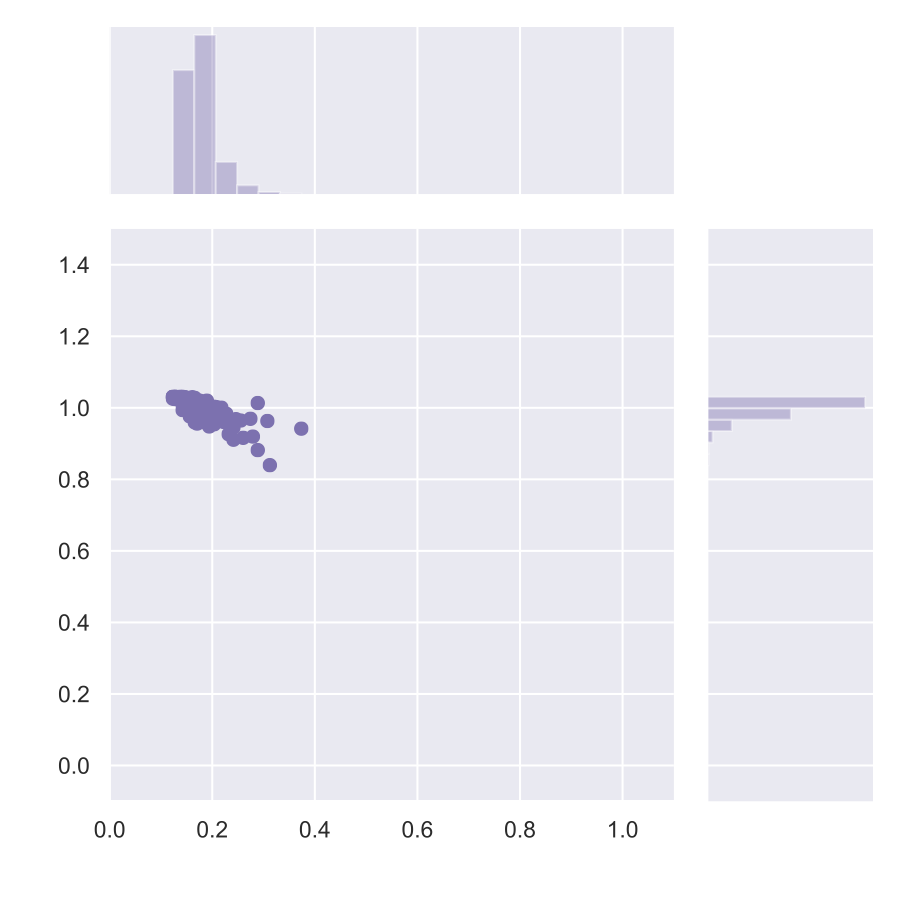}
  \caption{\textit{walker-stand}}
\end{subfigure}%

\leavevmode\smash{\makebox[0pt]{\hspace{.22\linewidth}% HORIZONTAL POSITION           
  \rotatebox[origin=l]{90}{\hspace{2.9em}% VERTICAL POSITION
    Performance (scaled)}%
}}\hspace{0pt plus 1filll}\null

\begin{center}
Inference costs (scaled)
\end{center}

\caption{Comparison of performance and cost. Each dot corresponds to a rollout of an episode. The \(y\)-axis is scaled so that the expert achieves 1 and a random policy achieves 0. The \(x\)-axis is also scaled such that only using $\pi_{\omega_{large}}$ throughout an episode corresponds to 1.}
\label{fig:perf_vs_cost}
% \vspace{-1em}
\end{figure*}

%\subsection{Behavior Analysis of the Decision Network}
\subsection{Qualitative Analysis of the Learned Behaviors}
\label{sec:analysis}

 We first illustrate a number of motivating timeline plots to qualitatively demonstrate that a control task can be handled by different sub-policies $\pi_{\omega}$ during different circumstances. 

\textbf{\textit{Swimmer-v3}.}
Fig.~\ref{fig:timeline_swimmer} illustrates the decisions of the master policy $\pi_{\Omega}$, where the interleavedly plotted white and yellow dots along the timeline correspond to the execution of sub-policies $\pi_{\omega_{small}}$ and $\pi_{\omega_{large}}$, respectively. 
In this task, a swimmer robot is expected to first perform a stroke and then maintain a proper posture so as to drift for a longer distance. It is observed that the model trained by our methodology tends to use $\pi_{\omega_{large}}$ while performing strokes and $\pi_{\omega_{small}}$ to maintain its posture between two strokes. One reason is that a successful stroke requires lots of delicate changes in each joint while holding a proper posture for drifting merely needs a few joint changes. Delicate changes in posture within a small time interval are difficult for $\pi_{\omega_{small}}$ since the outputs of it tend to be smooth over temporally neighboring states.
\textbf{\textit{MountainCarContinuous-v0}.} The objective of the car is to reach the flag at the top of the hill on the right-hand side. In order to reach the goal, the car has to accelerate forward and backward and then stop acceleration at the top. Fig.~\ref{fig:timeline_mcar} shows that $\pi_{\omega_{large}}$ is used for adjusting the acceleration and $\pi_{\omega_{small}}$ is only selected when acceleration is not required.

\textbf{\textit{FetchPickAndPlace-v1}.} The goal of the robotic arm is to move the black object to a target position (i.e., the red ball in Fig.~\ref{fig:timeline_fpap}). In Fig.~\ref{fig:timeline_fpap}, it can be observed that the agent trained by our methodology learns to use $\pi_{\omega_{small}}$ to approach the object, and then switch to $\pi_{\omega_{large}}$ to fetch and move it to the target location. One rationale for this observation is that fetching and moving an object entails fine-grained control of the clipper. The need for fine-grained control inhibits $\pi_{\omega_{small}}$ from being selected by $\pi_{\Omega}$ to fetch and move objects. In contrast, there is no need for fine-grained control for approaching objects. As a result, $\pi_{\omega_{small}}$ is mostly chosen when the arm is approaching the object to reduce the costs.

% The driving force of the car and the elevation of the hill are designed such that the car is unable to reach the flag by driving directly toward it. 
% The car has to climb the hill on the other side first to gain potential energy and then drives toward the flag. The process for the car to solve this task involves changes of its acceleration from a negative value to a positive value (i.e., drive backward at first then forward). Fig.~\ref{fig:timeline_mcar} shows that $\pi_{\Omega}$ uses $\pi_{\omega_{large}}$ while changing the acceleration, and uses $\pi_{\omega_{small}}$ to maintain its acceleration.

%% \textbf{\textit{Cartpole-swingup}.} In this task, an unactuated pole is attached to a cart. The objective of the cart is to swing up and balance the pole as long as possible. Fig.~\ref{fig:timeline}(c) shows that $\pi_\Omega$ uses $\pi_{\omega_{large}}$ to increase the rotation speed of the pole, which involves fast changes in the applied force, and then utilizes $\pi_{\omega_{small}}$ to decrease the rotation speed of the pole, which only requires a near-constant force. Once the pole reaches a steady upright angle, $\pi_{\omega_{small}}$ is then used by $\pi_\Omega$ to maintain the pose of the pole. 
%Actions with positive values correspond to increasing the clockwise rotation speed, and negative values correspond to increasing the counterclockwise rotation speed

\textbf{\textit{Walker-stand}.} The goal of the walker is to stand up and maintain an upright torso. Fig.~\ref{fig:timeline_walker_stand} shows that for circumstances when the forces applied change quickly, $\pi_{\omega_{large}}$ is used. For circumstances where the forces applied change slowly, $\pi_{\omega_{small}}$ is used. After the walker reaches a balanced posture, it utilizes $\pi_{\omega_{small}}$ to maintain the posture afterwards.

% The process for the walker to stand up involves the alternate use of $\pi_{\omega_{large}}$ and $\pi_{\omega_{small}}$. For circumstances where the forces applied change quickly, $\pi_{\omega_{large}}$ is used. For circumstances where the forces applied change slowly, $\pi_{\omega_{small}}$ is used. The alteration of the two sub-policies are depicted as the yellow and white dots in Fig.~\ref{fig:timeline_walker_stand}. After the walker reaches a balanced pose, it uses $\pi_{\omega_{small}}$ to maintain the pose afterwards. 

To summarize the above findings, $\pi_{\omega_{large}}$ is selected when find-grained controls (i.e., tweaking actions within a small time interval) are necessary, and $\pi_{\omega_{small}}$ is chosen otherwise.
\begin{table*}[t]
\centering
\caption{A summary of the performances of $\pi_{{S-only}}$, $\pi_{{L-only}}$, and our method (denoted as `\textit{Ours}') evaluated over 200 test episodes, along with the averaged percentages of $\pi_{\omega_{large}}$ being used by our method during an episode, as well as the averaged percentages of reduction in FLOPs when comparing \textit{Ours} (including the FLOPs from  $\pi_{\Omega}$ and the sub-policies) against $\pi_{{L-only}}$.}

% {Best score from 5 different seeds. \textcolor{red}{mention that they are calculated from 200 episodes. Mention that FLOPs means per inference?}

% \footnotesize
% \begin{center}
\resizebox{\linewidth}{!}{
 \begin{tabular}{c|cc|ccc} 
 \toprule
 Environment & $\pi_{{S-only}}$ & $\pi_{{L-only}}$ & \textbf{\textit{Ours}} & \textbf{\% using $\pi_{\omega_{large}}$} & \textbf{\% Total FLOPs reduction} \\ [0.5ex] 
 \midrule
 \textit{MountainCarContinuous-v0} & $-11.6\pm0.1$ & $93.6\pm0.1$ & $93.5\pm0.1$ & $44.5\%\pm5.7\%$ & $49.0\%\pm5.3\%$ \\
 \textit{Swimmer-v3} & $35.5\pm7.7$ & $84.1\pm18.0$ & $108.8\pm24.9$ & $54.9\%\pm9.5\%$ & $44.6\%\pm8.3\%$ \\ 
 \textit{Ant-v3} & $1,690.4\pm1,244.3$ & $3,927.2\pm1,602.8$ & $3,564.8\pm1,548.6$ & $53.9\%\pm5.5\%$ & $39.3\%\pm7.5\%$ \\
 \textit{FetchPickAndPlace-v1} & $0.351\pm0.477$ & $0.980\pm0.140$ & $0.935\pm0.255$ & $46.5\%\pm3.0\%$ & $46.4\%\pm2.8\%$ \\
 \textit{walker-stand} & $330.0\pm12.2$ & $977.7\pm22.1$ & $967.2\pm16.4$ & $5.7\%\pm1.1\%$ & $82.3\%\pm0.9\%$ \\ 
 \textit{finger-spin} & $32.9\pm36.9$ & $978.0\pm32.4$ & $871.2\pm24.0$ & $55.2\%\pm19.7\%$ & $37.5\%\pm17.8\%$ \\
 \bottomrule
\end{tabular}
}
% \end{center}
\label{tab:perf_best}
\end{table*}

% \subsection{Statistics of the Performance and Cost for the Proposed Methodology}
\subsection{Performance and Cost Reduction}
\label{sec:cost_vs_perf}
In this section, we compare the performance and the cost of our method with typical RL methods described in Section~\ref{sec:experimental_setup}. Table~\ref{tab:perf_best} summarizes the performances corresponding to $\pi_{{S-only}}$, $\pi_{{L-only}}$, and our method (denoted as `\textit{Ours}') in the second, third, and fourth columns, respectively. Table~\ref{tab:perf_best} also summarizes the averaged percentages of $\pi_{\omega_{large}}$ being used by our method during an episode, as well as the averaged percentages of reduction in FLOPs when comparing \textit{Ours} (including the FLOPs from the master policy $\pi_{\Omega}$ as well as the two sub-policies) against the $\pi_{{L-only}}$ baseline. % The evaluation results of the rest control task environments are offered in the supplementary material. 

It can be seen that in Table~\ref{tab:perf_best}, the average performance of \textit{Ours} are comparable to $\pi_{L-only}$ and significantly higher than $\pi_{S-only}$. It can also be observed that our method does switch between $\pi_{\omega_{small}}$ and $\pi_{\omega_{large}}$ to control the agent, and thus reduce the total cost required for solving the tasks.

To take a closer look into the behavior of the agent within an episode, Fig.~\ref{fig:perf_vs_cost} illustrates the performances and costs of our methodology over 200 episodes during evaluation for three control tasks. Each dot plotted in Fig.~\ref{fig:perf_vs_cost} corresponds to the evaluation result of \textit{Ours} in an episode, where the cost of each dot is divided by the cost of $\pi_{{L-only}}$. The performance of each dot is also scaled such that $[0, 1]$ corresponds to the averaged performances of a random policy and $\pi_{{L-only}}$. Please note that the scaled costs of our methodology may exceed one since the inference costs of  $\pi_{\Omega}$ are considered in our statistics as well. Histograms corresponding to the performances and costs of the data points are provided on the right-hand side and the top side of each figure, respectively. 
For \textit{Swimmer-v3}, it is observed that our methodology is able to reduce about half of the FLOPs when compared against $\pi_{{L-only}}$. Although few data points correspond to only half of the averaged performance of $\pi_{{L-only}}$, most of the data points are comparable and even superior to that. 
For \textit{FetchPickAndPlace-v1}, it can be observed that the dots distribute evenly along the line (y=1.0), which means that the agent can solve the tasks in the majority of episodes while the induced costs vary largely across episodes. This phenomena is mainly caused by the broadly varying starting positions in different episodes. When the object is close the arm, the cost is near 1.0 since $\pi_{\omega_{large}}$ is used in the majority of time in an episode, as the result shown in Section~\ref{sec:analysis}.
For \textit{walker-stand}, our method learns to use $\pi_{\omega_{large}}$ in the early stages to control the walker to stand up. After that, the agent only uses $\pi_{\omega_{small}}$ to slightly adjust its joints to maintain the posture of the walker. Therefore, a significant amount of inference costs can be saved in this task, causing the data points to concentrate on the top-left corner of the figure. These examples therefore validate that our cost-aware methodology is able to provide sufficient performances while reducing the inference costs required for completing the tasks. % Please note that additional examples, figures, and tables are provided in the supplementary material.

\subsection{Analysis of the Performance and the FLOPs per Inference}
\label{sec:baseline}

\begin{table*}[t]
\caption{
An analysis of the performances and FLOPs per inference (denoted as FLOPs/Inf) for our method and the baselines. The network sizes of $\pi_{{fit}}$ and the student networks of the two policy distillation baselines are configured such that their FLOPs/Inf are approximately the same as the averaged FLOPs/Inf of \textit{Ours} (denoted as Avg-FLOPs/Inf). In \textit{MountainCarContinuous-v0}, $\pi_{{L-only}}$, \textit{Ours}, and $\pi_{{fit}}$ are trained for 100k timesteps. In other control tasks, they are trained for 2M timesteps. BC and GAIL require additional expert trajectories generated by the trained model $\pi_{{L-only}}$, which consists of 25 trajectories with 50 state-action pairs for each trajectory, as adopted in~\cite{gail}. Note that the numerical results presented in this table correspond to the score of best model selected from 5 training runs.}

\label{tab:il_baselines}
\footnotesize
\begin{center}
\resizebox{\linewidth}{!}{

\begin{tabular}{c|cc|cc|cccc} 
  \toprule
  Environment & $\pi_{{L-only}}$ & FLOPs/Inf & \textit{Ours} & Avg-FLOPs/Inf & $\pi_{{fit}}$ & GAIL & BC & FLOPs/Inf \\ [0.5ex] 
  \midrule
  \textit{MountainCarContinuous-v0} & $93.6\pm0.1$ & 8,707 & $93.5\pm0.1$ & $4,440\pm177$ & $90.5\pm0.1$ & $-99.9\pm0.0$ & $93.3\pm0.1$ & 4,603 \\
  \textit{Swimmer-v3} & $84.1\pm10.3$ & 137,219 & \textbf{$108.8\pm15.4$} & $76,019\pm9,122$ & $66.2\pm10.1$ & $63.2\pm9.8$ & $59.7\pm13.8$ & 76,763 \\ 
  \textit{Ant-v3} & $3,927.2\pm524.0$ & 196,099 & \textbf{$3,564.8\pm724.7$} & $119,032\pm14,284$ & $2,553.0\pm511.7$ & $-15.6\pm101.0$ & $1,373.7\pm490.2$ & 119,451 \\
  \textit{FetchPickAndPlace-v1} & $0.980\pm0.140$ & 42,755 & $0.935\pm0.247$ & $22,917\pm2,521$ & $0.920\pm0.271$ & $0.078\pm0.268$ & $0.153\pm0.360$ & 23,223 \\
  \textit{walker-stand} & $977.7\pm20.2$ & 12,803 & \textbf{$967.2\pm16.3$} & $2,266\pm159$ & $819.5\pm14.5$ & $596.6\pm33.9$ & $159.1\pm22.6$ & 2,397 \\ 
  \textit{finger-spin} & $978.0\pm33.0$ & 9,859 & \textbf{$871.2\pm28.5$} & $6,162\pm739$ & $848.1\pm27.0$ & $536.8\pm22.4$ & $7.6\pm19.5$ & 6,303 \\
  \bottomrule
  \end{tabular}

}
\end{center}
% \vspace{-1em}
\end{table*}

\begin{table*}[t]
  \caption{
  Comparison of the proposed methodology with and without using the cost term $c_{\omega}$.
  % Comparison of the proposed methodology with and without policy costs subtracted from rewards.
  }\label{tab:ablation_no_cost}
  \centering
  \tiny
  % \resizebox{\linewidth}{!}{
      \begin{tabular}{ *{5}{c} }
        \toprule
        & \multicolumn{2}{c}{\textbf{With the cost term $c_{\omega}$}} &\multicolumn{2}{c}{\textbf{Without the cost term $c_{\omega}$}} \\
        \cmidrule(lr){2-3}\cmidrule(lr){4-5}
        \raisebox{\dimexpr1.25\normalbaselineskip-.5\height}[0pt][0pt]{\begin{tabular}{@{}c@{}}
          Environment
        \end{tabular}} & Performance & \% using $\pi_{\omega_{large}}$  & Performance & \% using $\pi_{\omega_{large}}$ \\
        \midrule
        \textit{MountainCarContinuous-v0} & $35.5\pm48.9$ & $50.4\%\pm5.5\%$ & $66.3\pm40.6$ & $59.0\%\pm20.5\%$ \\
        \textit{Swimmer-v3} & $98.9\pm23.2$ & $65.2\%\pm15.0\%$ & $71.5\pm33.3$ & $99.5\%\pm1.1\%$ \\
        \textit{Ant-v3} & $2,558.8\pm1140.0$ & $47.8\%\pm15.0\%$ & $2,625.8\pm728.6$ & $80.9\%\pm39.2\%$ \\
        \textit{FetchPickAndPlace-v1} & $0.822\pm0.103$ & $51.3\%\pm13.8\%$ & $0.785\pm0.175$ & $44.2\%\pm16.0\%$ \\
        \textit{walker-stand} & $943.8\pm23.3$ & $19.7\%\pm9.8\%$ & $961.8\pm10.0$ & $100.0\%\pm0.0\%$\\
        \textit{finger-spin} & $829.6\pm54.2$ & $38.6\%\pm14.9\%$ & $907.4\pm29.7$ & $100.0\%\pm0.0\%$\\ 
        \bottomrule
      \end{tabular}
 % }
  % \vspace{-1em}
\end{table*}
\begin{table*}[!tbh]
  \caption{
  Comparison of our methodology with and without a shared experience replay buffer.
  }
  \renewcommand{\arraystretch}{1.2}
  \label{tab:ablation_no_share_buffer}
  \centering
  \tiny
  % \resizebox{\textwidth}{!}{
      \begin{tabular}{ *{5}{c} }
        \toprule
        & \multicolumn{2}{c}{\textbf{With shared $\mathcal{Z_\omega}$}} &\multicolumn{2}{c}{\textbf{Without shared $\mathcal{Z_\omega}$}} \\
        \cmidrule(lr){2-3}\cmidrule(lr){4-5}
        \raisebox{\dimexpr1.25\normalbaselineskip-.5\height}[0pt][0pt]{\begin{tabular}{@{}c@{}}
          Environment
        \end{tabular}} & Performance & \% using $\pi_{\omega_{large}}$  & Performance & \% using $\pi_{\omega_{large}}$ \\
        \midrule
        \textit{MountainCarContinuous-v0} & $35.5\pm48.9$ & $50.4\%\pm5.5\%$ & $0.0\pm0.0$ & $59.8\%\pm54.6\%$ \\
        \textit{Swimmer-v3} & $98.9\pm23.2$ & $65.2\%\pm15.0\%$ & $61.1\pm23.6$ & $50.8\%\pm45.8\%$ \\
        \textit{Ant-v3} & $2,558.8\pm1,140.0$ & $47.8\%\pm15.0\%$ & $1,270.1\pm1,331.4$ & $23.8\%\pm42.6\%$ \\
        \textit{FetchPickAndPlace-v1} & $0.822\pm0.103$ & $51.3\%\pm13.8\%$ & $0.377\pm0.128$ & $63.0\%\pm17.4\%$ \\
        \textit{Walker-stand} & $943.8\pm23.3$ & $19.7\%\pm9.8\%$ & $910.1\pm65.9$ & $25.1\%\pm31.1\%$\\
        \textit{Finger-spin} & $829.6\pm54.2$ & $38.6\%\pm14.9\%$ & $865.9\pm41.8$ & $70.7\%\pm20.6\%$\\ 
        \bottomrule
      \end{tabular}
  % }
%   \vspace{2em}
\end{table*}

We compare the performances of the proposed methodology and the baselines discussed in Section~\ref{subsubsed::baselines}, as well as their FLOPs per inference (denoted as FLOPs/Inf). The FLOPs/Inf for $\pi_{{L-only}}$, \textit{Ours}, and the student networks of the baselines, as well as their corresponding highest performances achieved are summarized in Table~\ref{tab:il_baselines}. For a fair comparison, the sizes of the student networks of the distillation baselines are configured such that their FLOPs/Inf (the last column of Table~\ref{tab:il_baselines}) are approximately the same as the averaged FLOPs/Inf of \textit{Ours} (the Avg-FLOPs/Inf column in Table~\ref{tab:il_baselines}, including the FLOPs contributed by both the master policy $\pi_{\Omega}$ and the sub-policies). As a reference, we additionally train a policy $\pi_{{fit}}$ using SAC from scratch based on the same DNN size as the student networks of the distillation baselines. For distillation baselines, both of them employ the pre-trained $\pi_{{L-only}}$ as their teacher networks. Then, the student networks are trained using the data sampled from the trajectories generated by the teacher networks, where 50 consecutive state-action pairs are sampled from each of the generated 25 trajectories, as those adopted in~\cite{gail}. 

The results show that for the environments in Table~\ref{tab:il_baselines}, \textit{Ours} deliver comparable performances to the $\pi_{{L-only}}$ baseline and outperforms the distillation baselines, under similar levels of FLOPs/Inf. From the perspective of data samples used, the distillation baselines consume more data samples (including the data samples required for training both the teacher and the student networks) than those required by \textit{Ours}, which is trained from scratch without the need of data samples from a pre-trained teacher network. The relatively lower performances of the distillation baselines are probably due to the smaller sizes of the networks compared to their teacher networks $\pi_{{L-only}}$, since the performances delivered by $\pi_{{fit}}$ are also lower than the corresponding performances of \textit{Ours}. The results thus suggest that our method is able to reduce inference costs while maintaining sufficient performances. % Additional results of the student networks trained with longer periods or with more data samples for the distillation baselines are provided in the supplementary material. 

% \textcolor{red}{Our model is trained from scratch, without the need of data samples from a pre-trained expert network. Explain more details for the training data of the student networks.}

%Even if these student networks are trained longer or with more data samples, they are still unable to achieve the same performances as \textit{Ours}. \textcolor{purple}{These are additionally discussed in detail in the supplementary material}.

% We compare our method to GAIL in this section. We use the experts with the highest scores as teachers to generate trajectories for a GAIL policy to learn. The scores of the expert and the learned GAIL are shown in \ref{tab:il_baselines}. Except for the \textit{Hopper-v3} environment, our method does better than the GAIL baseline. We find that GAIL is not able to learn the behavior of teachers in all three \textit{Fetch} tasks, but our method can achieve a relatively high score in these tasks.

% \vspace{-0.5em}
\subsection{Ablation Study}
\label{sec:ablation}

\noindent\textbf{Effectiveness of the cost term.} We compare the evaluation results of our models trained with and without using the loss term in Table~\ref{tab:ablation_no_cost}. When the cost term is removed, the main factor that affects the decisions of  $\pi_{\Omega}$ is its belief in how good each sub-policy can achieve. Since $\pi_{\omega_{large}}$ is able to obtain high scores on its own, it is observed that $\pi_{\Omega}$ prefers to select $\pi_{\omega_{large}}$. In contrast, incorporating the cost term decreases the percentages of using $\pi_{\omega_{large}}$ substantially, while still allowing our model to offer satisfying performances.

\noindent \textbf{Effectiveness of shared experience replay buffer}. We compare the results of our models with and without the shared buffer across sub-policies $\pi_\omega$ in Table~\ref{tab:ablation_no_share_buffer}. For tasks except \textit{finger-spin}, the scores of the models without a shared $\mathcal{Z_\omega}$ are lower than those with a shared $\mathcal{Z_\omega}$. The lower scores of the models are due to reduced data samples for each sub-policies, since the transitions are not shared across the replay buffers. We also observed that some of the model trained without a shared $\mathcal{Z_\omega}$ is prone to use one of its sub-policies for the majority of time, instead of using both interleavedly. We believe that this is caused by unbalanced training samples for the two sub-policies. Namely, the relatively worse sub-policy is less likely to obtain sufficient data samples to improve its performance. While this problem can be solved by using training algorithms with improved exploration such as \cite{a3c}, we simply share $\mathcal{Z_\omega}$ among $\pi_\omega$ to address this issue. The models trained with a shared $\mathcal{Z_\omega}$ have lower variances in the choice of the two sub-policies (i.e., the third column of Table~\ref{tab:ablation_no_share_buffer}), and can exhibit more stable behaviors for $\pi_\Omega$.

\section{Conclusion}
\label{sec::conclusion}

We proposed a methodology for performing cost-aware control based on an asymmetric architecture. Our methodology uses a master policy to select between a large sub-policy network and a small sub-policy network. The master policy is trained to take inference costs into its consideration, such that the two sub-policies are used alternately and cooperatively to complete the task. The proposed methodology is validated in a wide set of control environments and the quantitative and qualitative results presented in this paper show that the proposed methodology provides sufficient performances while reducing the inference costs required. The comparison of the proposed methodology and the baseline methods indicated that the proposed methodology is able to deliver comparable performance to the $\pi_{{L-only}}$ baseline, while requiring less training data than the knowledge distillation baselines. % We further provided an ablation study to validate the effectiveness of the cost term in our loss function design.
% We additionally offered a wide set of experimental results and analyses in the supplementary material.

\section*{Acknowledgement}
This work was supported by the Ministry of Science and Technology (MOST) in Taiwan under grant nos. MOST 110-2636-E-007-010 (Young Scholar Fellowship Program) and MOST 110-2634-F-007-019. The authors acknowledge the financial support from MediaTek Inc., Taiwan. The authors would also like to acknowledge the donation of the GPUs from NVIDIA Corporation and NVIDIA AI Technology Center (NVAITC) used in this research work.

\bibliography{IEEEabrv,reference}
\end{document}

\makeatletter
\renewcommand{\@maketitle}{%
    \newpage\null\vskip2em%
    \begin{center}%
        \let\footnote\thanks{\LARGE\textbf{\@title}\par}%
        \vskip1.0em{\large\lineskip.5em\begin{tabular}[t]{c}\@author\end{tabular}\par }%
    \end{center}%
    \par}
\makeatother

\maketitle

% \renewcommand{\contentsname}{Table of Contents}
% \tableofcontents

% \clearpage

%%% Customization %%%

% Add prefix "S" to numbering
% \renewcommand{\thealgorithm}{S\arabic{algorithm}}
% \renewcommand{\thefigure}{S\arabic{figure}}
% \renewcommand{\thesection}{S\arabic{section}}
% \renewcommand{\thetable}{S\arabic{table}}
% \renewcommand{\theequation}{S\arabic{equation}}

%%% Main Text %%%
\section{Additional Background Material}

In this section, we provide addition background material. We first describe the basic concepts of deep Q-network (DQN)~\citep{dqn} and soft actor-critic (SAC)~\citep{sac}, which are used for training our master policy and sub-policies, respectively. We then briefly explain the concepts of hindsight experience replay (HER)~\citep{her} and Boltzmann exploration for DQN~\citep{boltzmann_done_right}, which are utilized in our experiments.  Finally, we provide some additional information that is related to hierarchical reinforcement learning (HRL).

\subsection{Deep Q-Learning (DQN)}

DQN~\citep{dqn} is a model-free RL algorithm based on deep neural networks (DNNs) for estimating the Q-function over high-dimensional state space. DQN is parameterized by a set of network weights $\phi$, which can be updated by a variety of RL algorithms. Given a policy $\pi$ and state-action pairs $(s,a)$, DQN incrementally updates its set of parameters $\phi$ such that $Q(s,a, \phi)$ approximates the optimal Q-function $Q^{*}$. The parameters $\phi$ are trained to minimize the loss function $L(\phi)$ iteratively using samples $(s, a, r, s^{\prime})$ drawn from an experience replay buffer $\mathcal{Z}$.  $L(\phi)$ is represented as the following:
\begin{equation} \label{eq::q_loss}
	L(\phi) = \mathbb{E}_{s,a,r,s^\prime \sim U(\mathcal{Z})}\big[(y - Q(s,a, \phi))^2\big],
\end{equation}
where $y = r + \gamma\max_{a^\prime} Q(s^\prime,a^\prime, \phi^{-})$, $r$ is the reward signal, $\gamma$ is the discount factor,  $(s',a')$ is the next state-action pair, $U(\mathcal{Z})$ is a uniform distribution over $\mathcal{Z}$, and $\phi^{-}$ represents the parameters of a target network.   The target network is the same as the online network parameterized by $\phi$, except that its parameters $\phi^{-}$ are updated by the online network periodically at constant intervals. Both the experience replay buffer and the target network enhance stability of the learning process dramatically.

%DQN~\citep{dqn} is a model-free approach to RL based on DNNs for estimating the Q-function over high-dimensional and complex state space. DQN is parameterized by a set of network weights $\phi$, which can be updated by a variety of RL algorithms. 
%To approximate the optimal Q-function given a policy $\pi$ and state-action pairs $(s,a)$, DQN incrementally updates its set of parameters $\phi$ such that $Q^{*}(s,a) \approx Q(s,a, \phi)$. The parameters $\phi$ are learned by gradient descent which iteratively minimizes the loss function $L(\phi)$ using samples $(s, a, r, s^{\prime})$ drawn from an experience replay memory $Z$.  $L(\phi)$ is expressed as:
%\begin{equation} \label{eq::q_loss}
%	L(\phi) = \mathbb{E}_{s,a,r,s^\prime \sim U(Z)}\big[(y - Q(s,a, \phi))^2\big]
%\end{equation}
%where $y = r + \gamma\max_{a^\prime} Q(s^\prime,a^\prime, \phi^{-})$, $(s',a')$ is the next state-action pair, $U(Z)$ is a uniform distribution over $Z$, and $\phi^{-}$ represents the parameters of the target network.   The target network is the same as the online network, except that its parameters $\phi^{-}$ are updated by the online network at predefined intervals.  Both the experience replay memory and the target network enhance stability of the learning process dramatically.

\subsection{Soft Actor-Critic (SAC)}
Soft actor-critic (SAC)~\citep{sac} is a deep RL algorithm which optimizes a stochastic policy in an off-policy manner. The key feature of SAC is the entropy regularization term in the loss function, which enables an agent to maximize the expected return and maintain the stochasticity of the actions during the training phase. SAC leans a policy $\pi_\theta$ and two Q-functions $Q_{\phi1}$ and $Q_{\phi2}$ at the same time. The two Q-functions are used to reduce the overestimation bias error from function approximation, as explained in the double Q-learning~\citep{double_q} paper. The target for the Q-functions is expressed as follows:
% Soft actor-critic (SAC)~\citep{sac} is a deep RL algorithm which optimizes a stochastic policy in an off-policy way. The key feature of SAC is the entropy regularization term in the loss function, which makes the model to maximize the expected return and maintain the stochasticity of action in the meanwhile. SAC leans a policy $\pi_\theta$ and two Q-function $Q_{\phi1}$ and $Q_{\phi2}$ at the same time. Two Q-functions are used to reduce the overestimation bias error from function approximation, as illustrated in the double Q-learning~\citep{double_q} paper. The target for the Q-functions can be expressed as:

\begin{equation} 
    \resizebox{\linewidth}{!}{
    	$y(r, s', d) = r + \gamma(1-d)\bigg( \min_{i=1,2} Q_{\phi_{targ,i}}(s', a', \phi_{targ,i}) - \alpha \log\pi_\theta(s', a') \bigg),
    	\label{eq::SAC_target}$
	}
\end{equation}
where $d$ is the terminal signal of an episode,  $\theta$ the parameters of the policy $\pi_{\theta}$, and $\alpha$ the entropy coefficient which controls the stochasticity of the policy. Based on Eq.~(\ref{eq::SAC_target}), the  Q-functions can be optimized to minimize the loss function $L(\phi)$ in Eq.~(\ref{eq::q_loss}). The policy $\pi_{\theta}$ can be updated to maximize:
% where $d$ is the terminal signal of an episode, and $\alpha$ is the entropy coefficient which controls the stochasticity of the policy. And then the Q-functions can be optimized to minimize the loss function $L(\phi)$ in Equation~\ref{eq::q_loss}. The policy can be updated to maximize
\begin{equation}
    \mathbb{E}_{s\sim\mathcal{Z}}\bigg( \min_{i=1,2} Q_{\phi_i}(s, a_\theta(s)) - \alpha\log\pi_\theta(a_\theta(s)|s) \bigg),
\end{equation}
where $\mathcal{Z}$ represents the replay buffer, and $a_\theta(s)$ denotes a sample from $\pi_\theta(.|s)$ which is differentiable with regard to the parameters $\theta$ of $\pi_{\theta}$ due to the use of the re-parameterization technique in SAC~\citep{sac}.

\subsection{Hindsight Experience Replay (HER)}
Consider an episode in a sparse reward setting environment with a state sequence $s_1,...,s_T$ and a goal $g\ne s_1,...,s_T$. This experience is not able to help the agent learn how to achieve the goal $g$, since no informative reward is acquired throughout this episode. In order to make the agent learn in such an environment, a more carefully designed reward function is required to guide the agent toward the goal. Instead of designing another reward function, HER~\citep{her} solves the above problem for an off-policy RL algorithm by replacing $g$ in the replay buffer with another pseudo goal, such that a large portion of the trajectories contain informative rewards which facilitate the learning of the agent. In addition, experience transitions with original goal $g$ is still available to the agent, such that the agent also learns to reach the true goals in the environment. There exist a number of strategies for choosing the pseudo goals for HER. In this paper, we use the `\textit{future}' strategy~\citep{her} for all of the experiments using HER, i.e., \textit{FetchPush-v1}, \textit{FetchPickAndPlace-v1}, and \textit{FetchPush-v1}, such that the pseudo goal is selected from the state achieved after the current timestep within the same episode.

% Consider an episode in a sparse reward setting environment with a state sequence $s_1,...,s_T$ and a goal $g\ne s_1,...,s_T$. This experience is not able to help the agent learn how to achieve the goal $g$, since no informative reward is acquired throughout this episode. In order to make the agent learn in such environment, a more carefully designed reward function is required to guide the agent toward the goal. Instead, HER~\citep{her} tries to solve this problem for an off-policy RL algorithm by replacing $g$ in the replay buffer with another pseudo goal, such that a large portion of the trajectories contain informative rewards which facilitate the learning of the agent. In addition, experience transitions with original goal $g$ is still available to the agent, such that the agent also learns to reach the true goals in the environment. There are a number of strategies to choose the pseudo goals for HER. We use `\textit{future}' strategy for all the experiments using HER, i.e., \textit{FetchPush-v1}, \textit{FetchPickAndPlace-v1} and \textit{FetchPush-v1}, such that the pseudo goal is selected from the states achieved after the current timestep in the same episode.

\subsection{Boltzmann Exploration for DQN}
Boltzmann exploration is another action-selection strategy for DQN to explore its action space besides $\epsilon$-greedy. Boltzmann exploration applies softmax function over the evaluations of Q-function for each action and take these values as the probability of choosing each action. The higher the Q-value is for an action, the more likely the action will be chosen. This approach gives more chance to sub-optimal actions than the $\epsilon$-greedy approach. A drawback of Boltzmann exploration is that the interpretation of the Q-values after applying the softmax function as the probability for choosing an action may not be the best choice to aid exploration, and may lead to sub-optimal behaviors of the model~\citep{boltzmann_done_right}. Instead of using Boltzmann exploration during the training phase, we use it during the evaluation phase to enable $\pi_\Omega$ to have chances to choose the relatively worse $\pi_\omega$. This might lead to sub-optimal performance, however, 
% the trained policy is still able to perform approximately as well as the optimal $\pi_\omega$, since 
a slight performance drop is acceptable in exchange of the reduction of computational costs. We apply Boltzmann exploration to several tasks in our experiments, including \textit{BipedalWalker-v3}, \textit{FetchPush-v1}, \textit{FetchSlide-v1}, \textit{FetchPickAndPlace-v1}, \textit{Hopper-stand}, \textit{Fish-swim}, and \textit{Reacher-easy}, where $\pi_\Omega$ tends to use one of its $\pi_\omega$ for an entire episode without the use of Boltzmann exploration.

    \begin{algorithm*}[t]
\SetAlgoLined
\SetKwInOut{Input}{input}\SetKwInOut{Output}{output}
\LinesNumbered
% \footnotesize
\small
\Input{total training steps \(T_{max}\), environment \(\mathcal{E}\)}
\DontPrintSemicolon
  
Initialize master policy~\(\pi_\Omega\) and sub-policies~\(\pi_{\omega_{small}}\) and \(\pi_{\omega_{large}}\)\;
Initialize global step counter and episode step counter: \(T\gets0\), \(t\gets0\)\;
Initialize experience replay buffer for \(\pi_\Omega\) and \(\pi_\omega\): \(\mathcal{Z}_\Omega\gets\{\}\), \(\mathcal{Z}_\omega\gets\{\}\)\;
Get initial state \(s\)\ from environment \(\mathcal{E}\)\;
\While{\(T<T_{max}\)}{
    \If(\Comment*[f]{select \(\pi_\omega\) every \(n_\omega\) timesteps}){\(t \bmod n_\omega == 0\)}{ 
        choose \(\omega\) according to \(\pi_\Omega(\omega|s)\)\;
        \(s_\Omega\gets s\)\;
        \(r_\Omega\gets0\)\
    }
    choose \(a\) according to \(\pi_\omega(a|s)\) \Comment*[r]{act according to the chosen \(\pi_\omega\)}
    take action \(a\) in \(s\), observe state \(s'\), reward \(r\), and terminal signal \(d\)\;
    \(r_\Omega\gets r_\Omega+r - \lambda c_{\omega}\)\Comment*[r]{penalize \(\pi_\Omega\) with weighted computational cost}
    \(t\gets t+1\)\;
    \;
    \If(\Comment*[f]{add transitions to replay buffer}){\(t \bmod n_\omega == 0 \text{ or } d == True\)}{
        \(\mathcal{Z}_\Omega\gets\mathcal{Z}_\Omega\cup\{(s_\Omega, \omega, r_\Omega, s', d)\}\)\;
    }
    \(\mathcal{Z}_\omega\gets\mathcal{Z}_\omega\cup\{(s, a, r, s', d)\}\)\;
    \;
    \If(\Comment*[f]{reset the environment when an episode ends}){d == True}{
        \(t\gets0\)\;
        reset environment and get a new state \(s\)\;
    }
    \;
    \If(\Comment*[f]{update \(\pi_\Omega\), \(\pi_{\omega_{small}}\) and \(\pi_{\omega_{large}}\)}){\(\mathcal{Z}_\Omega\) and \(\mathcal{Z}_\omega\) have enough samples}{
        update \(\pi_\Omega\) using sampled batches from \(\mathcal{Z}_\Omega\)\;
        update both \(\pi_{\omega_{small}}\) and \(\pi_{\omega_{large}}\) using sampled batches from \(\mathcal{Z}_\omega\)\;
    }
    \;
    \(T\gets T+1\)\;
    
}
\caption{The training procedure of the proposed cost-aware control methodology}
\label{algo}
\end{algorithm*}

\section{The Detailed Pseudo-Code of \\the Proposed Algorithm}

Algorithm~\ref{algo} summarizes the training procedure of  our methodology. In lines $6$-$9$, \(\pi_\Omega\) decides which \(\pi_\omega\) to use at a constant interval $n_{\omega}$. In lines $10$-$11$, the selected \(\pi_\omega\) determines an action and interact with $\mathcal{E}$. In line $12$, the reward for \(\pi_\Omega\) is derived by subtracting the weighted cost $\lambda c_\omega$ from $r_t$. Then, in lines $15$-$17$, experience transitions collected by \(\pi_\Omega\) and \(\pi_\omega\) are stored into the replay buffers $\mathcal{Z}_\Omega$ and $\mathcal{Z}_\omega$, respectively. Finally, in lines $23$-$25$, both \(\pi_\Omega\) and \(\pi_\omega\) are updated using sampled batches from $\mathcal{Z}_\Omega$ and $\mathcal{Z}_\omega$, respectively. The entire training procedure continues until the end of the horizon $T_{max}$.

    \begin{table}[!tb]
  \caption{
  The policy cost $c_{\omega_{small}}$,  $c_{\omega_{large}}$, as well as the policy cost coefficient $\lambda$ for each task.
  % The policy cost $c_{\omega_{small}}$ and $c_{\omega_{large}}$ as well as the choice of policy cost coefficient $\lambda$ for each task. In the last column, we also provide the ratio of policy cost to the episode return, denoted as $c_{total}/Score_{Avg}$, where $c_{total}=\mathbb{E}_{\omega\sim\pi_\Omega}[\sum_{t=0}^{T}\lambda c_{\omega_t}]$ and $Score_{Avg}$ is the average score achieved by our methodology.
  }
  \label{tab:policy_cost_coefficient}
  \centering
  \small
  \renewcommand{\arraystretch}{1.1}
  % \resizebox{\columnwidth}{!}{
      \begin{tabular}{c|ccc}
        \toprule
        Environment & $c_{\omega_{small}}$ & $c_{\omega_{large}}$ & $\lambda$ \\
        \midrule
        \textit{MountainCarContinuous-v0} & 1.0 & 44.7 & $1\mathrm{e}{-4}$\\
        \textit{BipedalWalker-v3} & 1.0 & 12.0 & $1\mathrm{e}{-4}$ \\
        \textit{HalfCheetah-v3} & 1.0 & 20.0 & $8\mathrm{e}{-2}$ \\
        \textit{Swimmer-v3} & 1.0 & 424.8 & $1\mathrm{e}{-4}$\\
        \textit{Ant-v3} & 1.0 & 8.0 & $1\mathrm{e}{-1}$ \\
        \textit{Walker2d-v3} & 1.0 & 12.3 & $3\mathrm{e}{-2}$ \\
        \textit{FetchPush-v1} & 1.0 & 17.5 & $8\mathrm{e}{-4}$ \\
        \textit{FetchSlide-v1} & 1.0 & 11.5 & $5\mathrm{e}{-5}$ \\
        \textit{FetchPickAndPlace-v1} & 1.0 & 9.4 & $2\mathrm{e}{-4}$ \\
        \textit{walker-stand} & 1.0 & 18.1 & $1\mathrm{e}{-2}$ \\
        \textit{finger-spin} & 1.0 & 29.1 & $1\mathrm{e}{-2}$ \\
        \textit{cartpole-swingup} & 1.0 & 37.4 & $3\mathrm{e}{-3}$ \\
        \textit{ball\_in\_cup-catch} & 1.0 & 30.1 & $1\mathrm{e}{-3}$\\
        \textit{hopper-stand} & 1.0 & 12.8 & $8\mathrm{e}{-4}$ \\
        \textit{fish-swim} & 1.0 & 220.0 & $1\mathrm{e}{-4}$ \\
        \textit{reacher-easy} & 1.0 & 32.6 & $2\mathrm{e}{-3}$ \\
        \bottomrule
      \end{tabular}
  % }
  % \vspace{-1em}
\end{table}

% \section{Additional Details of the Experimental Setup}
\section{Details of the Experimental Setup}

In this section, we provide details of our experimental setup, including the selection criteria of $c_{\omega}$ and $\lambda$, the network structures, as well as the hyperparameters used by our methodology and the baselines.

\subsection{Selection Criteria of the Policy Cost $c_{\omega}$ and the Coefficient $\lambda$}

\begin{figure}[t]
  \centering
  \includegraphics[width=\linewidth]{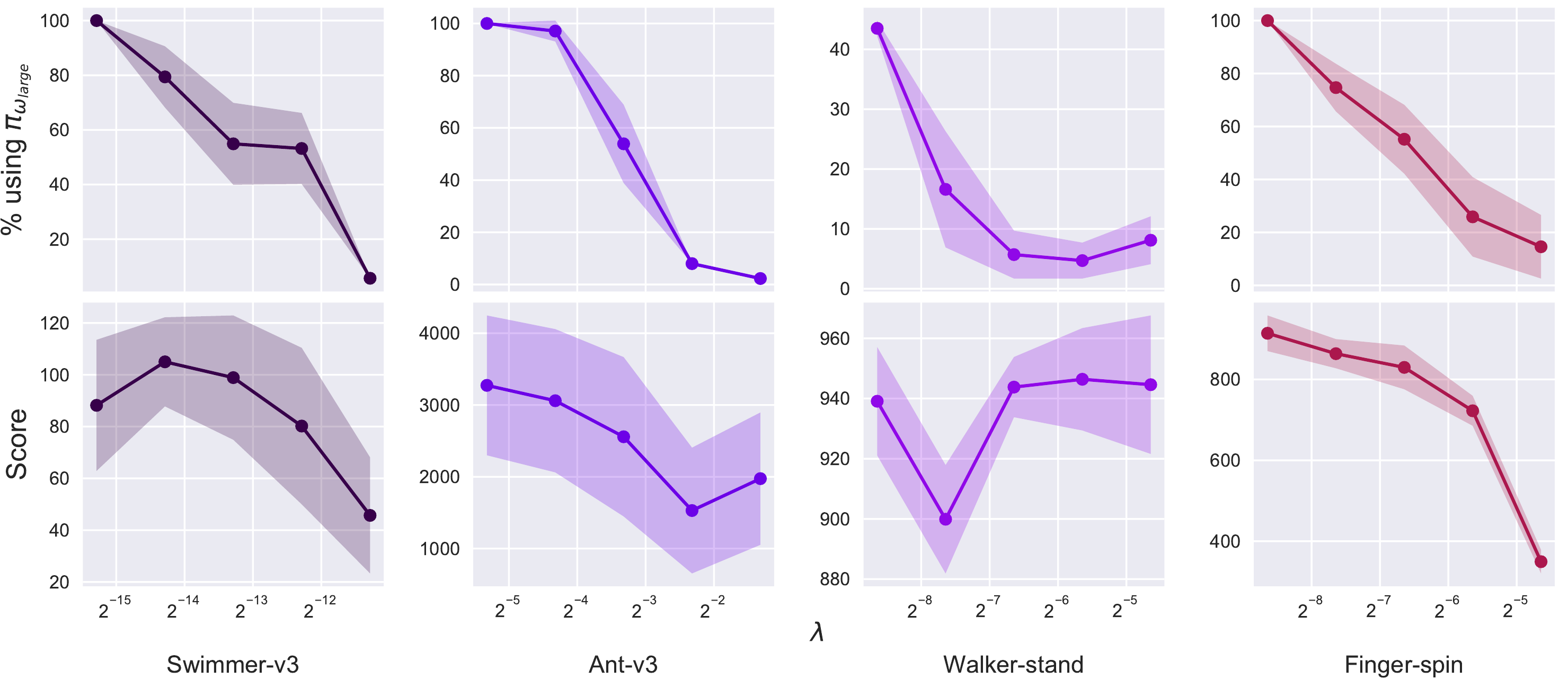}
  \caption{Performance of the models trained with different $\lambda$. The scores are averaged from 5 different random seeds. Each model trained with different random seed is evaluated over 200 episodes.}
  \label{fig:diff_lambda}
\end{figure}

We use the number of FLOPs of $\pi_{\omega_{small}}$ and $\pi_{\omega_{large}}$ divided by the number of FLOPs of $\pi_{\omega_{small}}$ as their policy costs $c_{\omega_{small}}$ and $c_{\omega_{large}}$, respectively, such that $c_{\omega_{small}}$ is equal to one. With regard to $\lambda$, from Fig.~\ref{fig:diff_lambda}, we observe that the relationship between $\lambda$ and the ratio of choosing $\pi_{\omega_{large}}$ is negatively correlated. In addition, the performances decline along with the reduced usage rate of $\pi_{\omega_{large}}$. We notice that there is often a range of $\lambda$ (ranging around the middle points in the figure), which allows us to develop candidate cost-efficient models that are potentially able to strike a balance between performance and usage rate of $\pi_{\omega_{large}}$. We then perform a hyperparameter search to find an appropriate $\lambda$, such that both $\pi_{\omega_{small}}$ and $\pi_{\omega_{large}}$ are used alternately within an episode, while allowing the agent to obtain high scores. Table~\ref{tab:policy_cost_coefficient} summarizes the values of $c_{\omega_{small}}$, $c_{\omega_{large}}$, and $\lambda$ used in each of the environments.

\subsection{Selection of the Master Policy Step Size $n_\omega$}
\begin{figure}[t]
  \centering
  \includegraphics[width=\linewidth]{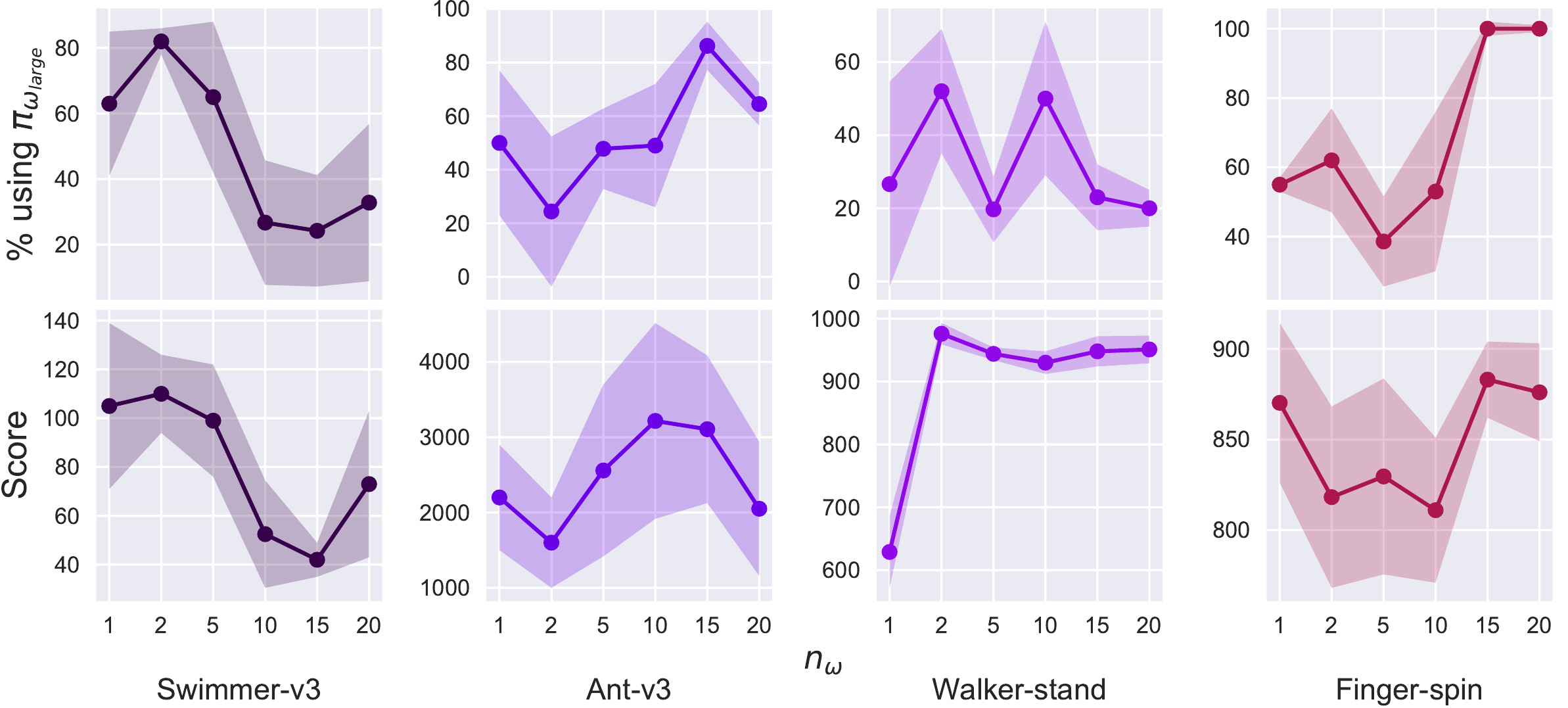}
  \caption{Performance of models trained with different $n_\omega$. The scores are averaged from 5 different random seeds. Each model trained with different random seed is evaluated over 200 episodes.}
  \label{fig:diff_nomega}
\end{figure}
It can be observed from Fig.~\ref{fig:diff_nomega} that there is no obvious correlation between $n_\omega$ and the performance. \textit{Swimmer-v3} performs well with smaller values of $n_\omega$. \textit{Ant-v3} performs well with $n_\omega$ equal to around $10$. On the other hand, \textit{walker-stand} performs well with larger values of $n_\omega$. Therefore, the choice of $n_\omega$ is relatively non-straightforward. We select the value of $n_\omega$ on account of two considerations: (1) $n_\omega$ should not be too small, or it will lead to increased master policy costs due to more frequent inferences of the master policy to decide which sub-policy to be used next; (2) $n_\omega$ should not be too large, otherwise the model will not be able to perform flexible switching between sub-policies. For instance, in the case of \textit{finger-spin}, a model with $n_\omega$ greater than $15$ uses $\pi_{\omega_{large}}$ throughout an episode while the usage rate of $\pi_{\omega_{small}}$ becomes almost zero. As a result, we set $n_\omega$ to $5$ for all of the experiments in this work as a compromise. 
An adaptive scheme of the step size $n_\omega$ may potentially enhance the overall performance and is left as a future research direction.

%However, an adaptive step size, i.e., $\pi_\Omega$ decides whether to switch to the other sub-policy on its own, may result in better overall performance. While this is not the main concern of this paper, we leave this as a possible future direction.

\subsection{Network Structure}
\label{network_structure}
% \textcolor{red}{Decribe the network structure of the proposed methodology, including the master policy and the sub-policies.}  \\

We implement both the master policy $\pi_\Omega$ and the sub-policies $\pi_\omega$ as multilayer perceptrons (MLPs) with two hidden layers of the same sizes. For all of the experiments, we choose the number of neurons per layer $n_{units}$ for $\pi_\Omega$ to be 32, such that the inference costs induced by $\pi_\Omega$ only account for a small portion of the overall costs, while giving $\pi_\Omega$ sufficient capability to assign task segments to different sub-policies. In order to reasonably determine the numbers of units per layer for $\pi_{\omega_{small}}$ and $\pi_{\omega_{large}}$, we first train a configuration with 512 units per layer for all selected tasks as our criterion policy $\pi_{criterion}$, and then train models with different number of units per layer, where $n_{units}\in\{8, 32, 64, 128, 196, 256\}$.  For each task, we set $n_{units}$ for $\pi_{\omega_{large}}$ such that its performance is above 90\% of the score of $\pi_{criterion}$. We adjust $n_{units}$ for $\pi_{\omega_{small}}$ such that 
its value is less than or equal to $1/4$ of $n_{units}$ for $\pi_{\omega_{large}}$, and the performance of $\pi_{\omega_{small}}$ is around or below $1/3$ of the score achieved by $\pi_{criterion}$. The exact values of $n_{units}$ for all the control tasks are listed in Table~\ref{tab:n_units}.

\begin{table*}[!tb]
  \caption{Number of units $n_{units}$ for $\pi_{\omega_{small}}$, $\pi_{\omega_{large}}$, and $\pi_{criterion}$, as well as their corresponding performances. The performances are averaged from the results trained with 5 different random seeds. }
  \label{tab:n_units}
  \centering
  \renewcommand{\arraystretch}{1.1}
  \footnotesize
  \resizebox{0.95\textwidth}{!}{
      \begin{tabular}{ *{7}{c} }
        \toprule
        & \multicolumn{2}{c}{\textbf{$\pi_{\omega_{small}}$}} 
        & \multicolumn{2}{c}{\textbf{$\pi_{\omega_{large}}$}} 
        & \multicolumn{2}{c}{\textbf{$\pi_{criterion}$}} \\
        \cmidrule(lr){2-3}\cmidrule(lr){4-5}\cmidrule(lr){6-7}
        \raisebox{\dimexpr1.25\normalbaselineskip-.5\height}[0pt][0pt]{\begin{tabular}{@{}c@{}}
          Environment
        \end{tabular}} & $n_{units}$ & Score  & $n_{units}$ & Score & $n_{units}$ & Score \\
        \midrule
        \textit{MountainCarContinuous-v0} & 8 & $-26.9\pm13.7$ & 64 & $87.8\pm4.8$ & 512 & $92.4\pm5.1$ \\
        \textit{BipedalWalker-v3} & 64 & $110.5\pm146.8$ & 256 & $297.5\pm12.0$ & 512 & $290.0\pm14.9$\\ 
        \textit{HalfCheetah-v3} & 8 & $1139.8\pm948.4$ & 64 & $8126.0\pm736.6$ & 512 & $7412.0\pm1442.7$\\
        \textit{Swimmer-v3} & 8 & $27.9\pm5.9$ & 256 & $66.4\pm13.7$ & 512 & $56.3\pm6.9$\\
        \textit{Ant-v3} & 64 & $1246.8\pm447.0$ & 256 & $2687.4\pm747.6$ & 512 & $2962.0\pm764.0$\\
        \textit{Walker2d-v3} & $64$ & $2282.6\pm290.6$ & 256 & $4156.6\pm148.0$ & 512 & $4202.0\pm343.8$ \\ 
        \textit{FetchPush-v1} & 8 & $0.094\pm0.009$ & 64 & $0.970\pm0.015$ & 512 & $0.972\pm0.011$\\ 
        \textit{FetchSlide-v1} & 64 & $0.100\pm0.068$ & 256 & $0.726\pm0.053$ & 512 & $0.666\pm0.063$\\
        \textit{FetchPickAndPlace-v1} & 32 & $0.212\pm0.149$ & 128 & $0.948\pm0.022$ & 512 & $0.970\pm0.010$ \\
        \textit{walker-stand} & 8 & $289.0\pm47.5$ & 64 & $891.3\pm143.2$ & 512 & $915.7\pm91.2$ \\
        \textit{finger-spin} & 8 & $47.1\pm75.8$ & 64 & $928.4\pm31.7$ & 512 & $862.6\pm88.1$\\
        \textit{cartpole-swingup} & 8 & $125.0\pm74.1$ & 64 & $819.8\pm24.1$ & 512 & $810.6\pm50.5$ \\
        \textit{ball\_in\_cup-catch} & 8 & $128.6\pm52.5$ & 64 & $970.0\pm13.2$ & 512 & $948.0\pm29.2$\\
        \textit{hopper-stand} & 64 & $223.8\pm165.9$ & 256 & $611.6\pm277.0$  & 512 & $640.1\pm230.5$\\
        \textit{fish-swim} & 8 & $74.6\pm2.1$ & 256 & $206.0\pm36.5$ & 512 & $193.1\pm44.5$  \\
        \textit{reacher-easy} & 8 & $166.5\pm83.2$ & 64 & $948.5\pm18.8$ & 512 & $954.0\pm37.2$ \\ 
        \bottomrule
      \end{tabular}
  }
%   \begin{tabularx}{\linewidth}{X}
%   *: When $n_{units}$ is reduced to $32$, the performance drops significantly. Thus, $n_{units}$ for $\pi_{\omega_{small}}$ is set to $64$.
%   \end{tabularx}
  
\end{table*}

The network structures for performing the experiments of the baselines also consist of two hidden layers with the same number of units, except that the number of units for both layers is chosen such that the FLOPs/Inf (number of FLOPs per inference) is approximately the same as the averaged FLOPs/Inf of \textit{Ours} calculated from 200 test episodes, as discussed in Section~5.3 of the main manuscript. We list the values of $n_{units}$ used for training $\pi_{fit}$, BC~\citep{bc_limitation}, and GAIL~\citep{gail} in Table~\ref{tab:baseline_units}.

% The network structures for running baseline experiments also consist of two hidden layers with the same number of units, except for that the number of units for both layers is chosen such that the $FLOPs/Inf$ (number of FLOPs per inference) is approximately the same as the $Avg-FLOPs/Inf$ (averaged number of FLOPs per inference) calculated from 200 episodes. We list $n_{units}$ used for training $\pi_{fit}$, GAIL and BC in Table~\ref{tab:baseline_units}.
\begin{table}[t]
  \caption{Number of units $n_{units}$ adopted for training the baselines $\pi_{fit}$, BC~\citep{bc_limitation}, and GAIL~\citep{gail}. Note that the corresponding results and discussions are given in Section 5.3 of the main manuscript.}
  \label{tab:baseline_units}
  \centering
  \renewcommand{\arraystretch}{1.1}
  \small
  % \resizebox{\columnwidth}{!}{
      \begin{tabular}{c|c}
        \toprule
        Environment & $n_{units}$\\
        \midrule
        \textit{MountainCarContinuous-v0} & 46 \\
        \textit{Swimmer-v3} & 190\\ 
        \textit{Ant-v3} & 189 \\
        \textit{FetchPickAndPlace-v1} & 90\\
        \textit{walker-stand} & 21\\
        \textit{finger-spin} & 50\\ 
        \bottomrule
      \end{tabular}
  % }
  %\vspace{-1em}
\end{table}

% \subsection{Units of the Hidden Layers}
% \input{supplementary/tables/n_units_of_hidden.tex}

\subsection{Hyperparameters for Training the Proposed Methodology and Baselines}

The hyperparameters used for training the proposed methodology are provided in Table~\ref{tab:ours_hyperparam}. On the other hand, the hyperparameters used for training the baseline methods are summarized in Table~\ref{tab:baseline_hyperparam}.
% \vspace{3em}
% \begin{table*}[t]
% \parbox[t]{.45\linewidth}{
% \input{supplementary/tables/ours_hyperparams.tex}
% }
% \hfill
% \parbox[t]{.45\linewidth}{
% \input{supplementary/tables/baseline_hyperparams.tex}
% }
% \end{table*}

%\subsection{Settings of the Environments}
%\textcolor{red}{Desccribe the environmental setups of the control tasks, if any.}

%\subsection{Other Hyperparameters (\textcolor{red}{Optional})}

%\subsection{Setup of the Unity Environments}
%\textcolor{red}{Describe the environment setup of the Unity tasks}
%We use Unity as our environment platform and utilize Unity Machine Learning Agents Toolkit(ML-Agent) to interact with the environment. In this environment,  the agent needs to simultaneously avoid obstacles and reaching the ultimate goal to complete the navigation task. Training environments consist of different scenarios, such as straight road, curved road or route with obstacles.

\section{Additional Experimental Results of the Proposed Methodology}

% In this section, we present both the additional qualitative and quantitative results for our methodology.

% (i.e., yellow and white dots in the bottom left sub-figure in Fig.~\ref{fig:timeline}(d). 

% \section{Analysis of $\pi_{\Omega}$}
% In Fig.~\ref{fig:perf_vs_cost}, we see from the regression lines that performance and costs have a positive correlation. The more a policy chooses \(\pi_{\omega_{large}}\), the better the performance. In most of the tasks, the model could save up to 50\% of costs with a little performance drop. However, for some of the tasks such as \textit{Swimmer} and \textit{FetchPickAndPlace}, the performance drop is not significant. We believe that it is due to the properties of the tasks. These tasks require different control complexity in different periods within an episode. For example, in \textit{Swimmer-v3} the swimmer only applies forces at a constant interval, since the inertia lasts for a while. By contrary, tasks such as \textit{HalfCheetah} and \textit{Walker2d} require sustained precision control. As a result, only little costs can be saved for such tasks. We show more concrete examples in the next section.

\subsection{Statistics of the Performance and Cost for the Proposed Methodology}
Following the discussions presented in Section~5.3 of the main manuscript, in this section, we provide the  results of the other control tasks. The performances, the percentages of using $\pi_{\omega_{large}}$, the percentages of the total FLOPs reduction, as well as the performances of $\pi_{S-only}$ and $\pi_{L-only}$ are listed in Table~\ref{tab:extra_experiments}. For columns 2-6, the number reported in each entry is an average of the results from five different random seeds, where the result corresponding to each seed is averaged over 200 episodes. For columns 7-9, the best  results of our proposed methodology are presented, which reveal that our methodology is able to balance the tradeoff between performance and computational costs.

% We provide more experimental results in this section. The performances, percents of using $\pi_{\omega_{large}}$ and percents of total FLOPs reduction, as well as the performances of $\pi_{S-only}$ and $\pi_{L-only}$ are listed in Table~\ref{tab:extra_experiments}. For columns 2-6, each entry is an average over training from five different random seeds, and each training is evaluated over 200 episodes. For columns 7-9, we list the best training results of our methodology, which find a great balance between performance and computational costs.

For tasks except for \textit{HalfCheetah-v3} and \textit{FetchSlide-v1}, Table~\ref{tab:extra_experiments} shows that our methodology results in only slight performance drops (when compared with $\pi_{{L-only}}$), while reducing a significant amount of computational costs for most of the tasks. For \textit{FetchSlide-v1}, the performance drop is primary due to the reduction in the usage of $\pi_{\omega_{large}}$. For \textit{HalfCheetah-v3}, our model tends to use either $\pi_{\omega_{small}}$ or $\pi_{\omega_{large}}$ for an entire episode, despite of the usage of either the fine-tuned policy cost coefficient $\lambda$ or Boltzmann exploration during the evaluation phase. A potential reason might be due to the fact that the control complexity required by the model is approximately the same for the entire episode. Thus, it is difficult for $\pi_{\Omega}$ to learn to switch between $\pi_{\omega_{small}}$ and $\pi_{\omega_{large}}$, leading to the performance drop.

\begin{figure}[ht]
% \noindent\hrulefill\par
\noindent\makebox[\textwidth][c]{%
\resizebox{!}{.55\textheight}{
\begin{minipage}{\textwidth}
\begin{subfigure}{.333\textwidth}
  \centering
  \includegraphics[width=\linewidth]{figures/perf_vs_cost/moutaincar_random.pdf}
  \caption{\textit{MountainCarContinuous-v0}}
\end{subfigure}%
\begin{subfigure}{.333\textwidth}
  \centering
  \includegraphics[width=\linewidth]{figures/perf_vs_cost/ant_random.pdf}
  \caption{\textit{Ant-v3}}
\end{subfigure}%
\begin{subfigure}{.333\textwidth}
  \centering
  \includegraphics[width=\linewidth]{figures/perf_vs_cost/finger_spin.pdf}
  \caption{\textit{finger-spin}}
\end{subfigure}%
\newline

\begin{subfigure}{.333\textwidth}
  \centering
  \includegraphics[width=\linewidth]{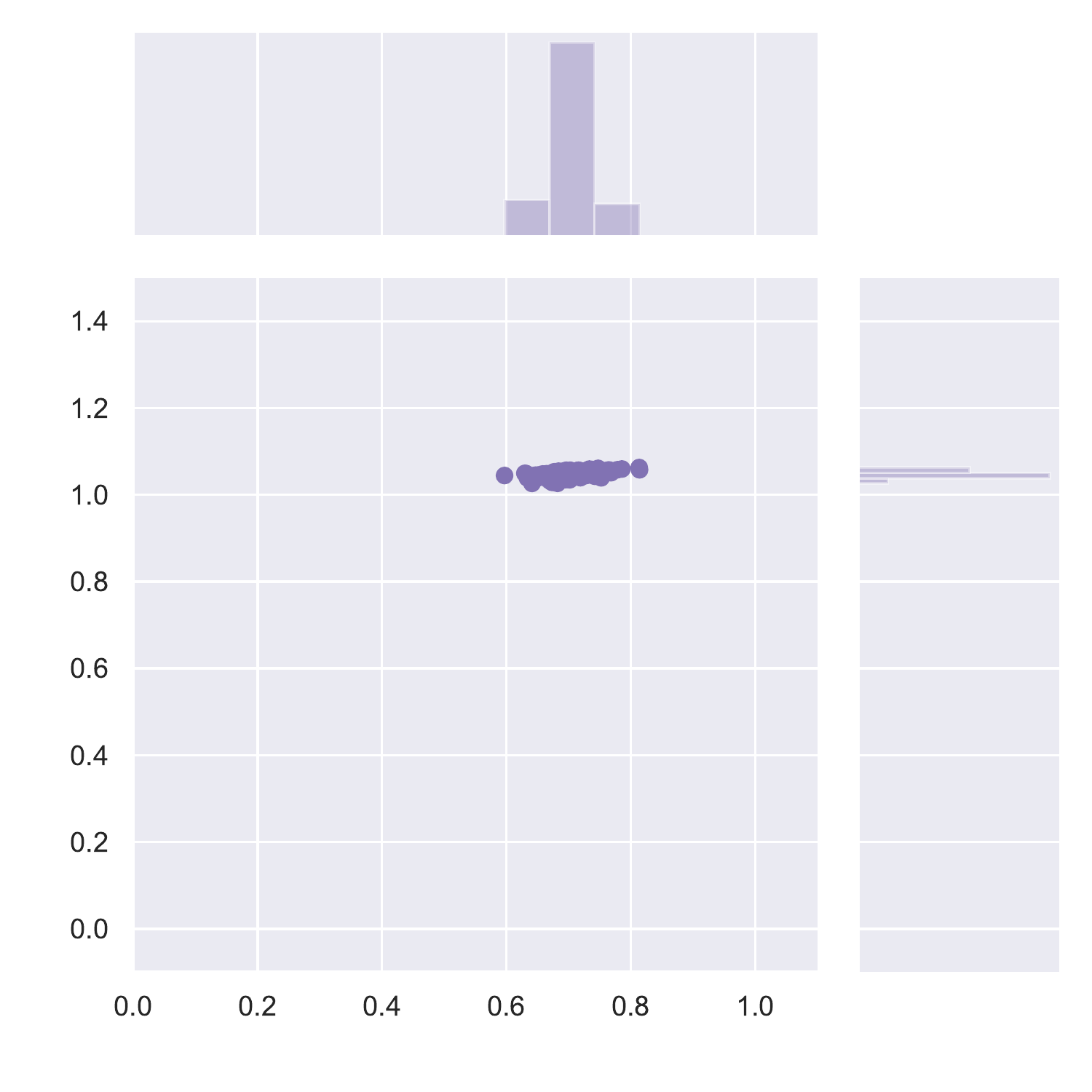}
  \caption{\textit{BipedalWalker-v3}}
\end{subfigure}%
\begin{subfigure}{.333\textwidth}
  \centering
  \includegraphics[width=\linewidth]{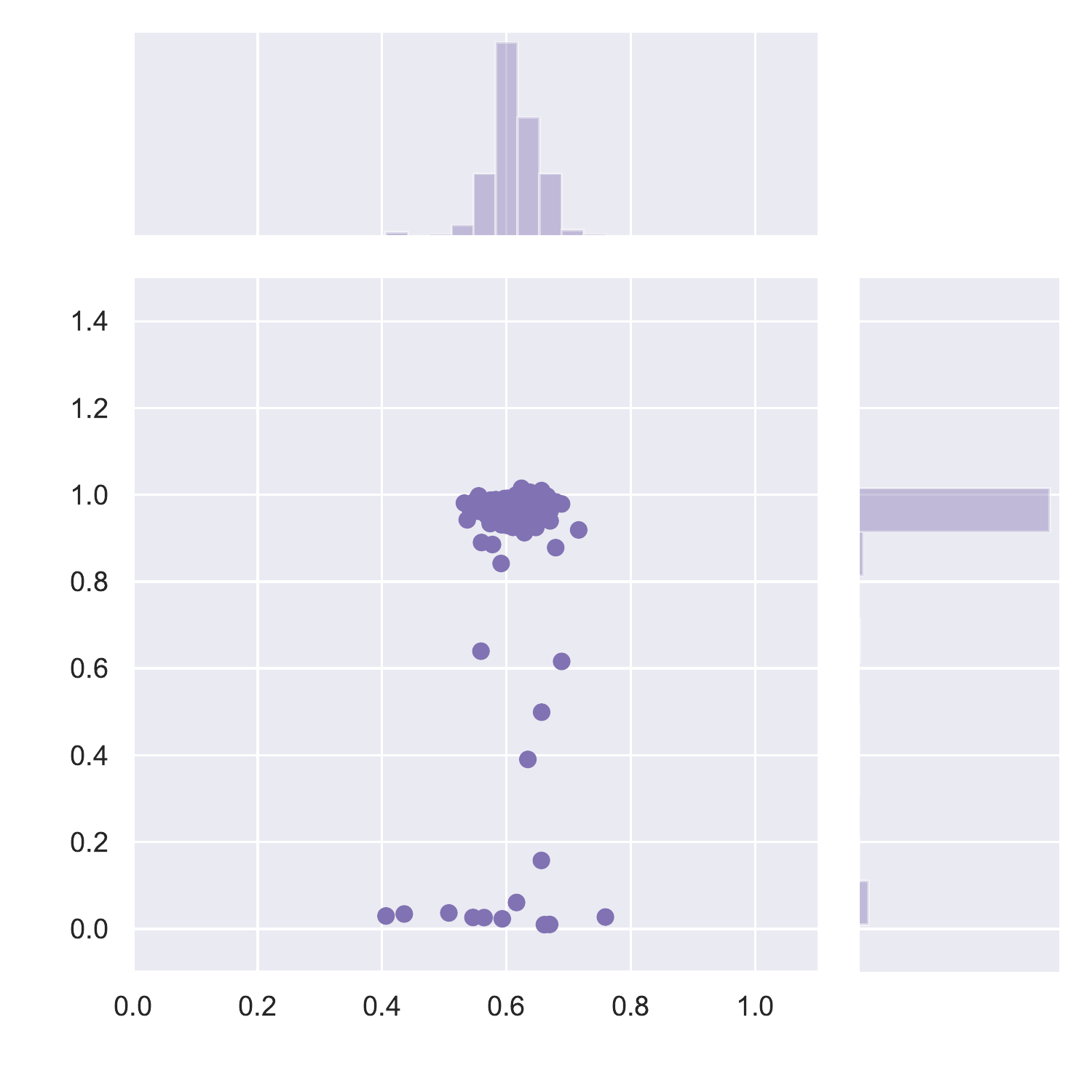}
  \caption{\textit{Walker-v3}}
\end{subfigure}%
\begin{subfigure}{.333\textwidth}
  \centering
  \includegraphics[width=\linewidth]{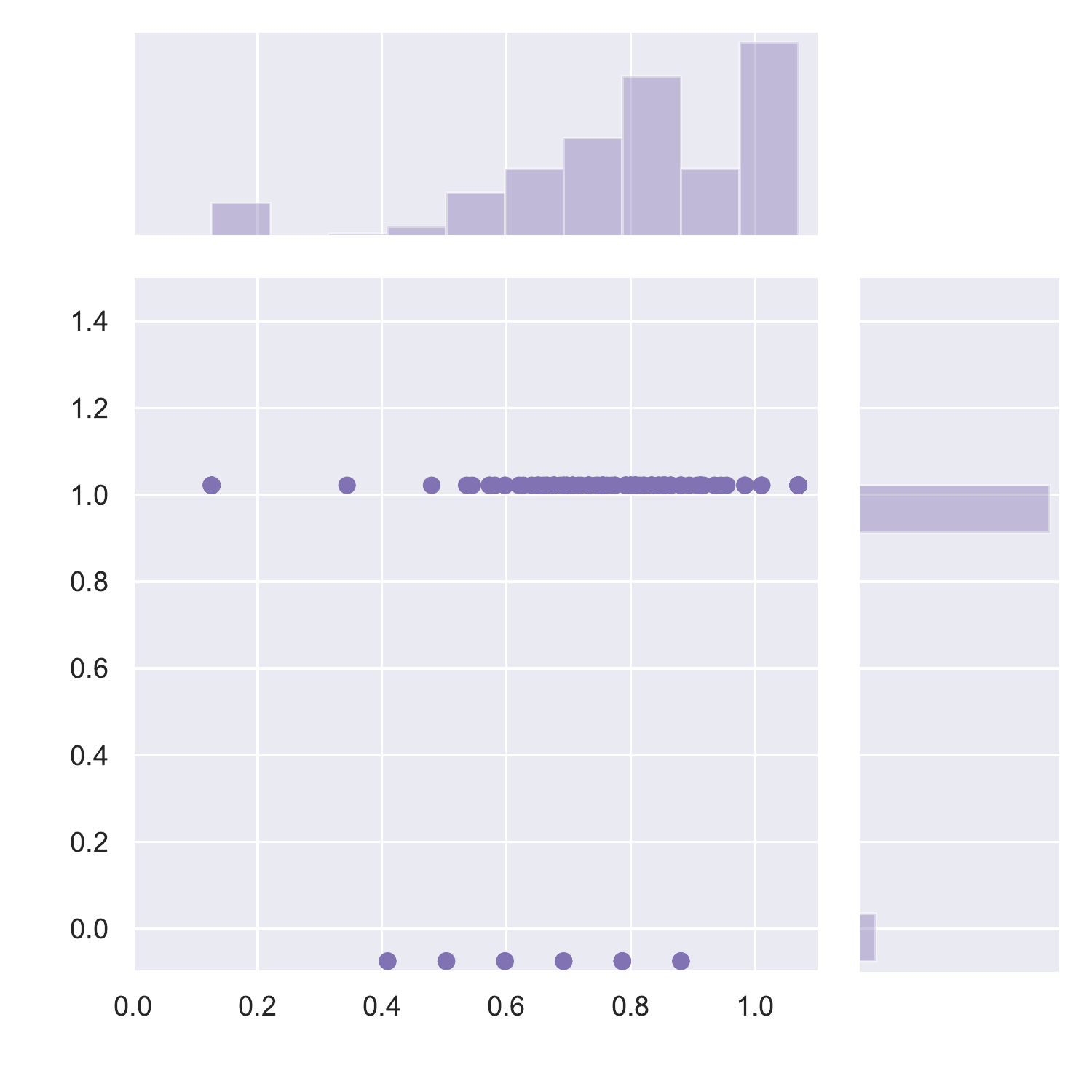}
  \caption{\textit{FetchPush-v1}}
\end{subfigure}%
\newline

\begin{subfigure}{.333\textwidth}
  \centering
  \includegraphics[width=\linewidth]{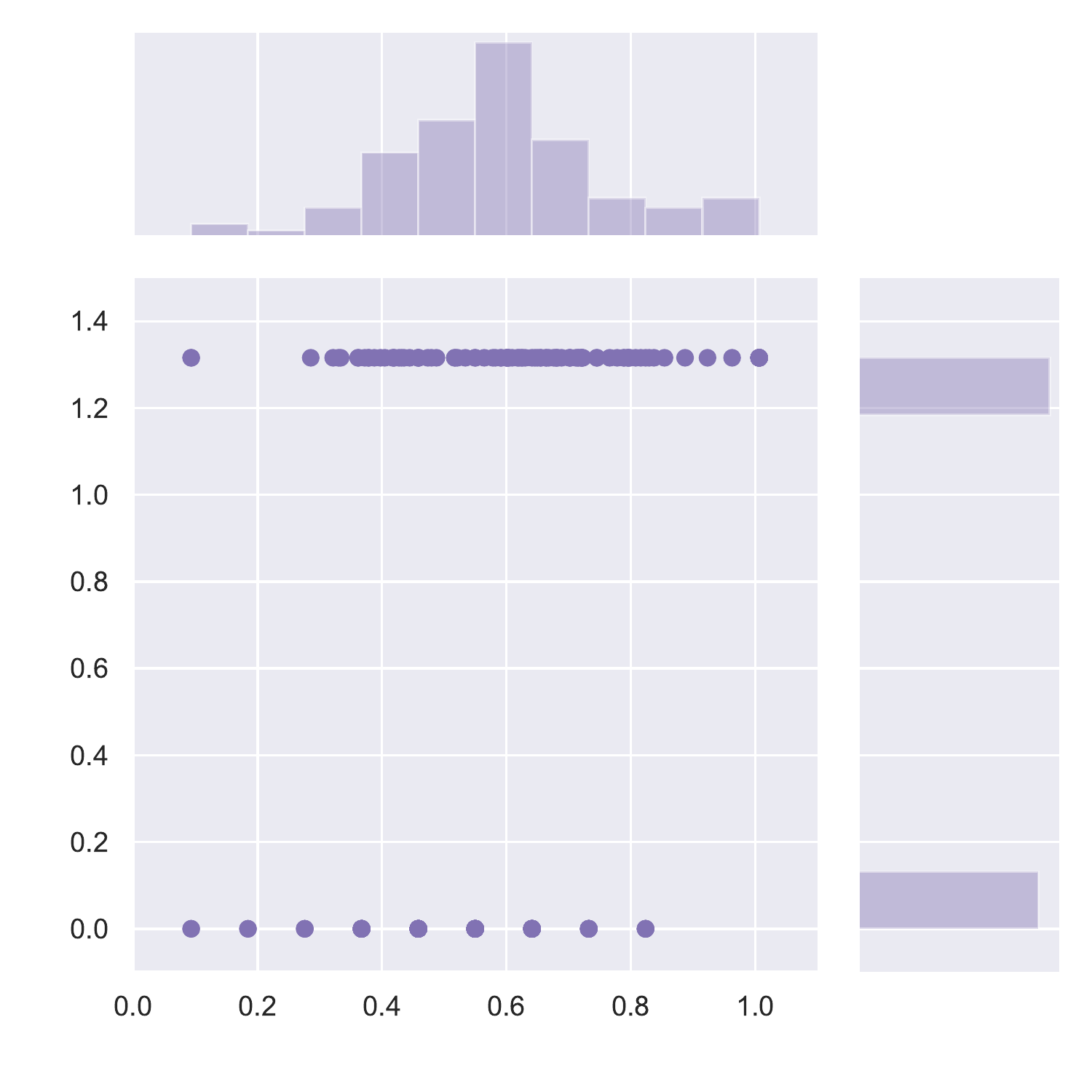}
  \caption{\textit{FetchSlide-v1}}
\end{subfigure}%
\begin{subfigure}{.333\textwidth}
  \centering
  \includegraphics[width=\linewidth]{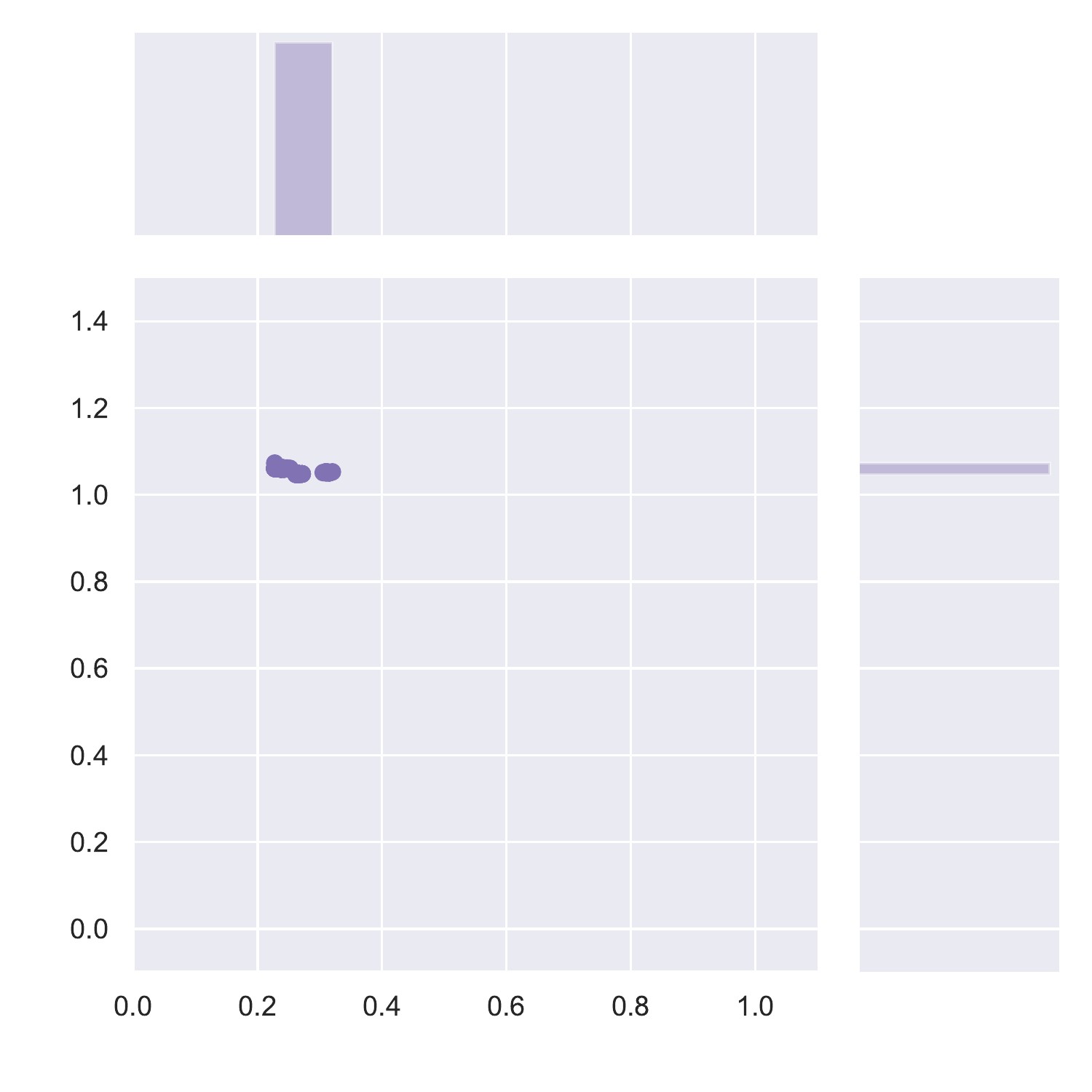}
  \caption{\textit{Cartpole-swingup}}
\end{subfigure}%
\begin{subfigure}{.333\textwidth}
  \centering
  \includegraphics[width=\linewidth]{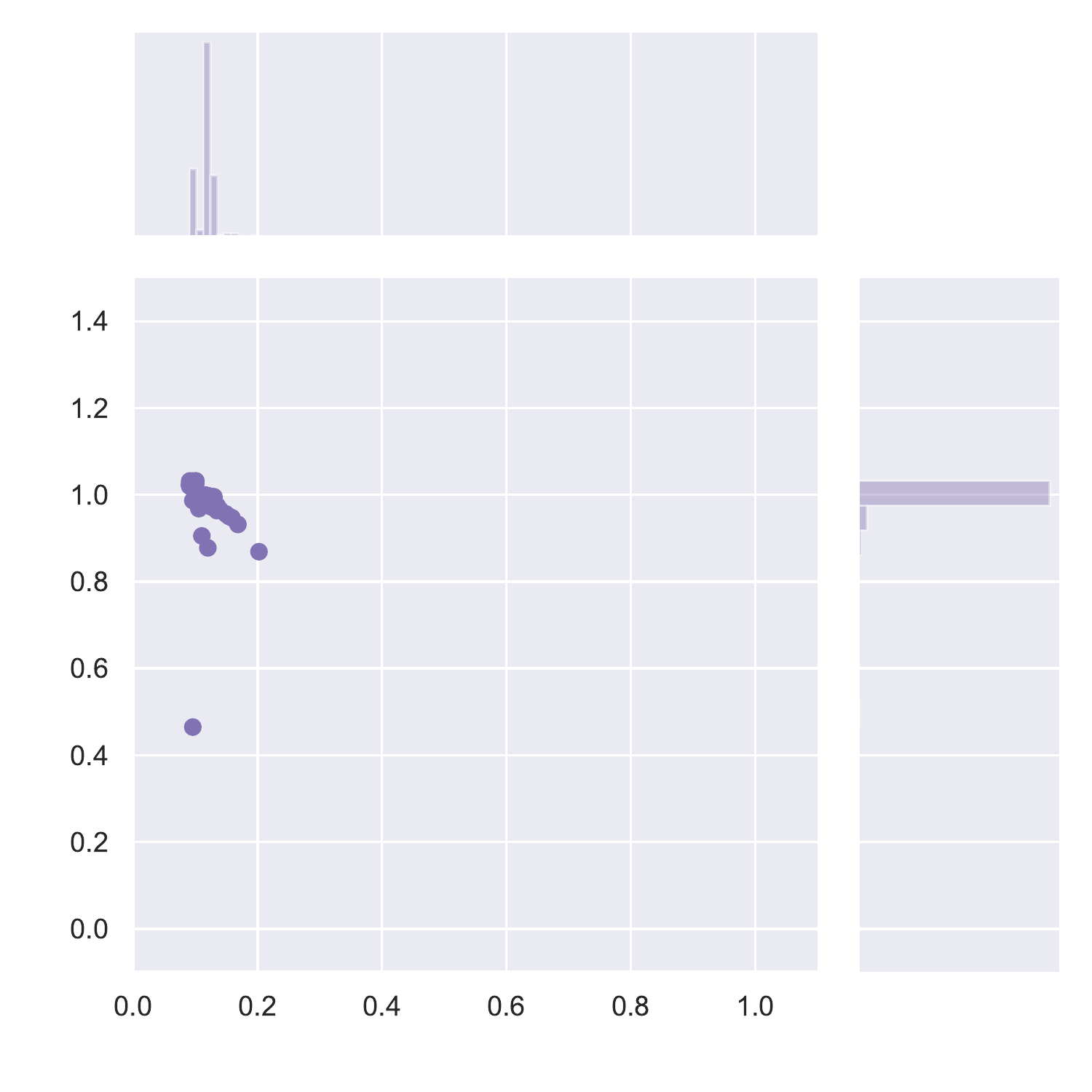}
  \caption{\textit{Ball\_in\_cup-catch}}
\end{subfigure}%
\newline

\begin{subfigure}{.333\textwidth}
  \centering
  \includegraphics[width=\linewidth]{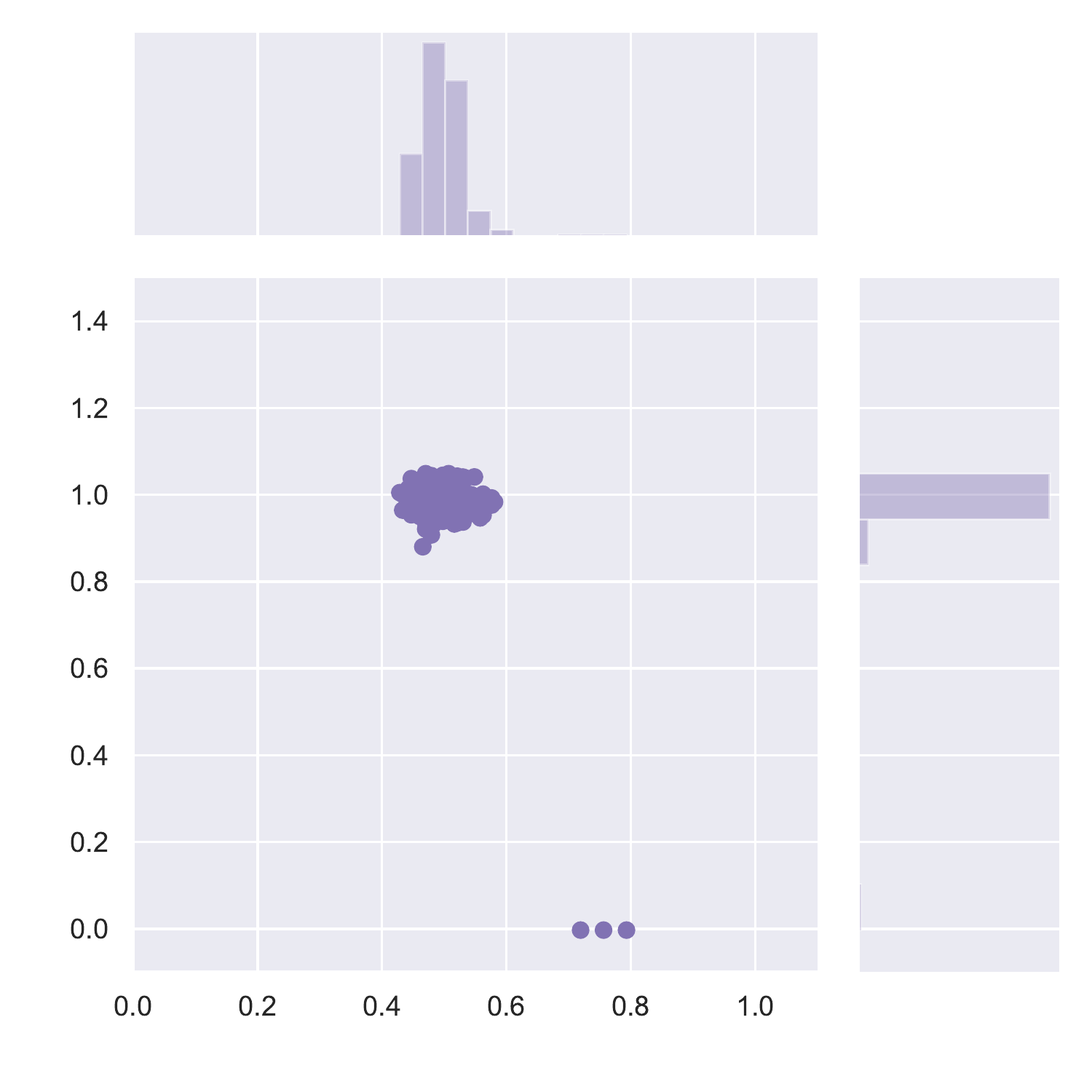}
  \caption{\textit{Hopper-stand}}
\end{subfigure}%
\begin{subfigure}{.333\textwidth}
  \centering
  \includegraphics[width=\linewidth]{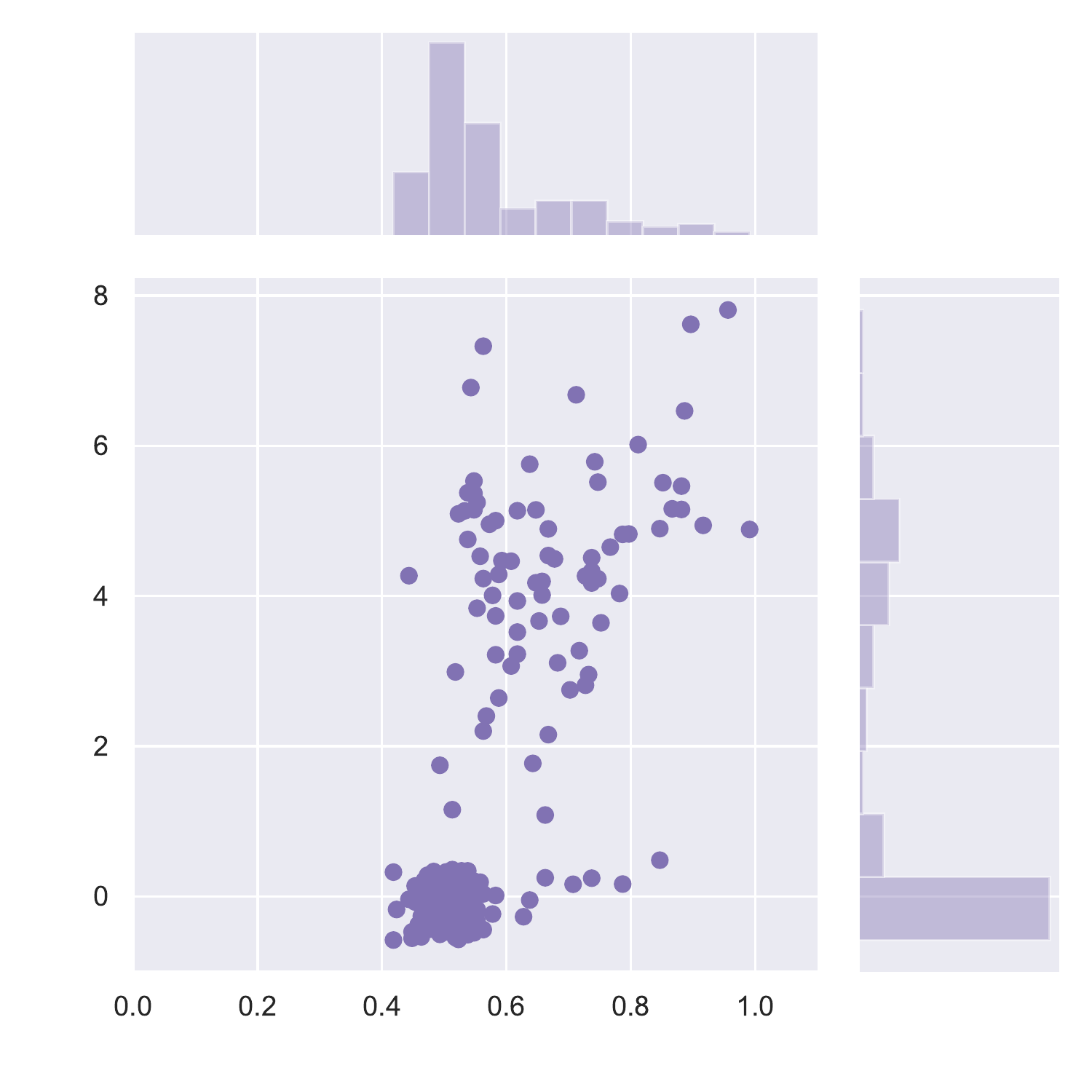}
  \caption{\textit{Fish-swim}}
\end{subfigure}%
\begin{subfigure}{.333\textwidth}
  \centering
  \includegraphics[width=\linewidth]{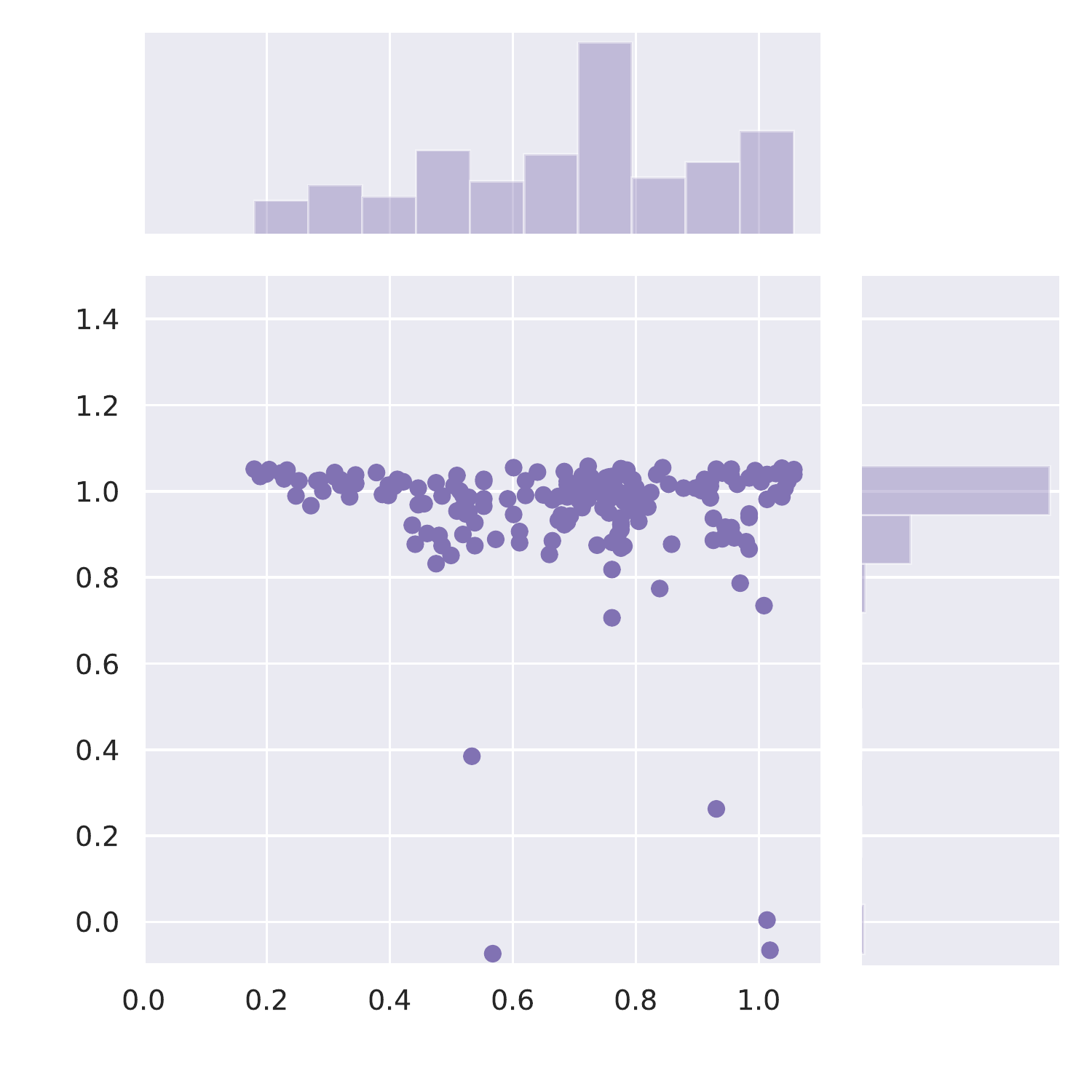}
  \caption{\textit{Reacher-easy}}
\end{subfigure}%
\newline
\leavevmode\smash{\makebox[0pt]{\hspace{0em}% HORIZONTAL POSITION           
  \rotatebox[origin=l]{90}{ \hspace{33.2em}% VERTICAL POSITION
    Performance (scaled)}%
}}\hspace{0pt plus 1filll}\null

\begin{center}
Computational costs (scaled)
\end{center}

\caption{Comparison of performance and cost. Each dot in the plots corresponds to a rollout of an episode. The \(y\)-axis is performance, scaled so that the expert achieves 1 and a random policy achieves 0. The \(x\)-axis is computational costs, scaled so that using large policy throughout an episode takes 1.}
\label{fig:sup_perf_vs_cost}
% \vspace{-1em}
\end{minipage}
}
}
\end{figure}

\vspace{0.5em}
The best models selected from the five training rounds listed in the last three columns of Table~\ref{tab:extra_experiments} further reveal the feasibility to train models to deliver comparable performances as those achieved by $\pi_{L-only}$, , while reducing a significant amount of computational costs. We do not show the standard deviation for the entries in the last three columns of Table~\ref{tab:extra_experiments}. Instead, the distribution plots with regard to the performances and the computational costs are illustrated in Fig.~\ref{fig:sup_perf_vs_cost}. For most of the tasks, the dots are concentrated on the upper part of the figures, indicating the stability of the performances of the models trained by our methodology. For the \textit{fish-swim} task, the model trained with the vanilla SAC has a high score variance for each episode, which leads to a high variance in the experimental result of our methodology inherently.  Nevertheless, the overall performance of our best model still outperforms  the best model trained with the vanilla SAC in the \textit{fish-swim} task in Table~\ref{tab:extra_experiments}.

\onecolumn
\begin{landscape}
\begin{table*}[t]
\parbox[t]{.65\textwidth}{
% \begin{table}[t]
\centering
\renewcommand{\arraystretch}{1.1}
\caption{
The detailed settings of the hyperparameters adopted by the master policy $\pi_\Omega$ and the sub-policies $\pi_\omega$ of our methodology.
}
\label{tab:ours_hyperparam}
\footnotesize
\resizebox{.8\columnwidth}{!}{
\begin{tabular}{lc|lc}
\toprule \toprule
\multicolumn{1}{c}{Hyperparameter}                                      & Value & \multicolumn{1}{c}{Hyperparameter}                                      & Value  \\ \toprule \toprule
\multicolumn{2}{c|}{\textbf{Master Policy $\pi_\Omega$}} & \multicolumn{2}{c}{\textbf{Sub-Policy $\pi_\omega$}}
            \\ \midrule
            RL algorithm & DQN & RL algorithm & SAC \\
            Learning rate & $1\mathrm{e}{-3}$ & Entropy coefficient $\alpha$ & Auto \\
    		Discount factor ($\gamma$) & $0.99$ & Learning rate of agent & $3\mathrm{e}{-4}$ \\
    		Replay buffer size & 50K & Discount factor ($\gamma$) & $0.99$ \\
    		Exploration fraction & 10\% & Replay buffer size & 1M \\ 
    		Update batch size & 32 & Update batch size & 256 \\
    		Double Q & True & Train frequency & 1 \\ 
    		Train frequency & 1 & Target soft update coefficient $\tau$ & 0.005 \\
    		Target network update interval & 500 & Target network update interval & 1 \\
            Optimization for the RL agent & Adam & Optimization for the RL agent & Adam \\
            Training timesteps & 2.5M & Training timesteps & 2.5M \\
            Nonlinearity & Tanh & Nonlinearity & Tanh \\
%             \toprule
% \multicolumn{2}{c}{\textbf{Sub-Policy $\pi_\omega$}}
%             \\ \midrule
%             RL algorithm & SAC \\
%             Entropy coefficient $\alpha$ & Auto \\
%     		Learning rate of agent & $3\mathrm{e}{-4}$ \\
%     		Discount factor ($\gamma$) & $0.99$ \\
%     		Replay buffer size & 1M \\
%     		Update batch size & 256 \\
%     		Train frequency & 1 \\
%     		Target soft update coefficient $\tau$ & 0.005 \\ 
%     		Target network update interval & 1 \\
%             Optimization for the RL agent & Adam \\
%             Training timesteps & 2.5M \\
%             Nonlinearity & Tanh \\
\bottomrule \bottomrule
\end{tabular}
}
% \end{table}
}
\hfill
\parbox[t]{.45\textwidth}{
% \begin{table}[t]
\centering
\caption{The detailed settings of the hyper-parameters adopted by the imitation learning baseline methods, including GAIL and BC.}
\label{tab:baseline_hyperparam}
\footnotesize
\begin{tabular}{lc}
\toprule \toprule
\multicolumn{1}{c}{Hyperparameter}                                      & Value   \\ \toprule \toprule
\multicolumn{2}{c}{\textbf{GAIL}} 
            \\ \midrule
            Adversary layers $n$  & 2 \\
            Adversary hidden size & 100 \\
            Adversary entropy coefficient & $1\mathrm{e}{-3}$ \\ 
            Update frequency for the policy & 3 \\
            Update frequency for the discriminator & 1 \\
            Value function step size & $3\mathrm{e}{-4}$ \\ 
            Training timesteps & 4M \\
            \toprule
\multicolumn{2}{c}{\textbf{BC}}
            \\ \midrule
    		Learning rate & $1\mathrm{e}{-4}$ \\
    		Training epoches & 100 \\
    		Optimizer & Adam \\
    		Update batch size & 64 \\
\bottomrule \bottomrule
\end{tabular}
% \end{table}
}
% \begin{table}[b]
\renewcommand{\arraystretch}{1.3}
\footnotesize
\small
\caption{Results of our methodology averaged from five different random seeds, where the result from each seed is averaged over 200 episodes.}

% {Best score from 5 different seeds. \textcolor{red}{mention that they are calculated from 200 episodes. Mention that FLOPs means per inference?}

% \begin{center}
% \resizebox{\columnwidth}{!}{
 \begin{tabular}{c|cc|ccc|ccc} 
 \toprule \toprule
 Environment & $\pi_{{S-only}}$ & $\pi_{{L-only}}$ & \textbf{\textit{\makecell{Ours \\(average)}}} & \textbf{\makecell{\% using \\$\pi_{\omega_{large}}$}} & \textbf{\makecell{\% Total FLOPs \\Reduction}} & \textbf{\textit{\makecell{Ours \\(best)}}} & \textbf{\makecell{\% using \\$\pi_{\omega_{large}}$}} & \textbf{ \makecell{\% Total FLOPs \\Reduction}} \\ [0.5ex] 
 \midrule
 \textit{BipedalWalker-v3} & $110.5\pm146.8$ & $297.5\pm12.0$ & $281.7\pm24.6$ & $68.5\%\pm5.1\%$ & $28.3\%\pm4.7\%$ & 309.1 & 67.6\% & 29.8\% \\
 \textit{HalfCheetah-v3} & $1,139.8\pm948.4$ & $8,126\pm736.6$ & $7,483.4\pm1,170.1$ & $99.4\%\pm0.9\%$ & $-6.0\%\pm0.9\%$ & 8,388.2 & 100.0\% & -6.6\%\\
 \textit{Walker2d-v3} & $2,282.6\pm290.6$ & $4,156.6\pm148.0$ & $4,059.7\pm340.7$ & $77.3\%\pm17.3\%$ & $20.3\%\pm15.9\%$ & 4,255.1 & 57.0\% & 39.0\% \\ 
 \textit{FetchPush-v1} & $0.094\pm0.009$ & $0.970\pm0.015$ & $0.826\pm0.119$ & $74.1\%\pm3.0\%$ & $17.5\%\pm2.8\%$ & 0.915 & 72.8\% & 18.7\%\\
 \textit{FetchSlide-v1} & $0.100\pm0.068$ & $0.726\pm0.053$ & $0.501\pm0.148$ & $57.6\%\pm27.8\%$ & $38.1\%\pm25.4\%$ & $0.595$ & 59.0\% &36.8\%\\
 \textit{Cartpole-swingup} & $125.0\pm74.1$ & $819.8\pm24.1$ & $817.5\pm44.7$ & $55.8\%\pm42.9\%$ & $37.5\%\pm41.7\%$ & 848.5 & 17.3\% & 75.0\%\\ 
 \textit{Ball\_in\_cup-catch} & $128.6\pm52.5$ & $970.0\pm13.2$ & $967.5\pm3.4$ & $3.0\%\pm0.6\%$ & $88.0\%\pm0.6\%$ & 964.5 & 2.5\% & 88.5\% \\
 \textit{Hopper-stand} & $223.8\pm165.9$ & $611.6\pm277.0$ & $534.9\pm322.2$ & $50.3\%\pm19.5\%$ & $45.3\%\pm17.9\%$ & 876.7 & 45.0\% & 50.2\%\\
 \textit{Fish-swim} & $74.6\pm2.1$ & $206.0\pm36.5$ & $201.5\pm59.6$ & $50.2\%\pm5.5\%$ & $51.0\%\pm7.5\%$ & 268.6 & 56.4\% & 42.8\% \\
 \textit{Reacher-easy} & $165.5\pm83.2$ & $948.5\pm18.8$ & $689.7\pm272.7$ & $50.4\%\pm28.5\%$ & $42.4\%\pm27.4\%$ & 931.7 & 60.0\% & 33.1\%\\
 \bottomrule \bottomrule
\end{tabular}
% }
% \end{center}
\label{tab:extra_experiments}
% \end{table}
\end{table*}

\end{landscape}

\twocolumn
\clearpage

% table and perf_cost plots

% \input{supplementary/tables/without_sharing_buffer.tex}

\section{Additional Ablation Studies}
\subsection{Sub-Policies with and without Separated Replay Buffers}
% \subsection{Sub-Policies using Separated Replay Buffers}
\begin{table*}[!tbh]
  \caption{
  Comparison of our methodology with and without a shared experience replay buffer.
  }
  \renewcommand{\arraystretch}{1.2}
  \label{tab:ablation_no_share_buffer}
  \centering
  \tiny
  % \resizebox{\textwidth}{!}{
      \begin{tabular}{ *{5}{c} }
        \toprule
        & \multicolumn{2}{c}{\textbf{With shared $\mathcal{Z_\omega}$}} &\multicolumn{2}{c}{\textbf{Without shared $\mathcal{Z_\omega}$}} \\
        \cmidrule(lr){2-3}\cmidrule(lr){4-5}
        \raisebox{\dimexpr1.25\normalbaselineskip-.5\height}[0pt][0pt]{\begin{tabular}{@{}c@{}}
          Environment
        \end{tabular}} & Performance & \% using $\pi_{\omega_{large}}$  & Performance & \% using $\pi_{\omega_{large}}$ \\
        \midrule
        \textit{MountainCarContinuous-v0} & $35.5\pm48.9$ & $50.4\%\pm5.5\%$ & $0.0\pm0.0$ & $59.8\%\pm54.6\%$ \\
        \textit{Swimmer-v3} & $98.9\pm23.2$ & $65.2\%\pm15.0\%$ & $61.1\pm23.6$ & $50.8\%\pm45.8\%$ \\
        \textit{Ant-v3} & $2,558.8\pm1,140.0$ & $47.8\%\pm15.0\%$ & $1,270.1\pm1,331.4$ & $23.8\%\pm42.6\%$ \\
        \textit{FetchPickAndPlace-v1} & $0.822\pm0.103$ & $51.3\%\pm13.8\%$ & $0.377\pm0.128$ & $63.0\%\pm17.4\%$ \\
        \textit{Walker-stand} & $943.8\pm23.3$ & $19.7\%\pm9.8\%$ & $910.1\pm65.9$ & $25.1\%\pm31.1\%$\\
        \textit{Finger-spin} & $829.6\pm54.2$ & $38.6\%\pm14.9\%$ & $865.9\pm41.8$ & $70.7\%\pm20.6\%$\\ 
        \bottomrule
      \end{tabular}
  % }
%   \vspace{2em}
\end{table*}

In this section, we validate the choice of using a shared experience replay buffer across sub-policies. We compare the evaluation results of our models with and without the shared buffer in Table~\ref{tab:ablation_no_share_buffer}. For tasks except \textit{finger-spin}, the scores of the models  with a shared buffer $\mathcal{Z_\omega}$ are higher than those without a shared $\mathcal{Z_\omega}$. The lower scores of the models trained without a shared $\mathcal{Z_\omega}$ are due to reduced data samples for training each sub-policies, since the transitions are not shared across the replay buffers. We also observed that some of the model trained without a shared $\mathcal{Z_\omega}$ is prone to use one of its sub-policies for the majority of time, instead of using both interleavedly. We believe that this is caused by the unbalanced training samples for the two sub-policies. In other words, the relatively worse sub-policy is less likely to obtain sufficient data samples to improve its performance. In contrast, the models trained with a shared $\mathcal{Z_\omega}$ have lower variances in the choice of the two sub-policies (e.g., please refer to the second column of Table~\ref{tab:ablation_no_share_buffer}), and are able to exhibit more stable behaviors for $\pi_\Omega$.  

% We validate the choice of choosing shared experience replay buffer across sub-policies in this section. We compare the evaluation results of our models with shared or non-shared buffer in Table~\ref{tab:ablation_no_share_buffer}. For tasks except \textit{finger-spin}, the scores of models with shared buffer $\mathcal{Z_\omega}$ is higher than that without shared buffer $\mathcal{Z_\omega}$. The lower scores of models without shared $\mathcal{Z_\omega}$ is due to reduced data samples for training each sub-policies, since the transitions are not shared across replay buffers. We also observed that some of the model trained without shared $\mathcal{Z_\omega}$ is prone to use one of its sub-policy for the majority of time, instead of using both interleavedly. We believe this is caused by unbalanced training for sub-policies, that is, the worse sub-policy never gets enough data samples to improve its performance. In contrast, the models trained with shared $\mathcal{Z_\omega}$ have lower variance in the choice of sub-policies (second column of Table~\ref{tab:ablation_no_share_buffer}) and have more stable behaviors for $\pi_\Omega$.  

\subsection{Comparison against an HRL Model with Two Large Sub-Policies}
\begin{table*}[!tb]
  \caption{Performance of model with two sub-policies, both of the same size as $\pi_{\omega_{large}}$. Each score is averaged from 5 independent training with different random seeds, and each trained model is evaluated over 200 test episodes.}
  \label{tab:all_large}
  \centering
  \renewcommand{\arraystretch}{1.1}
  \small
  \resizebox{\textwidth}{!}{
      \begin{tabular}{c|ccc}
        \toprule
        Environment & $\pi_{L-only}$ & \textit{Ours} & HRL with two $\pi_{\omega_{large}}$\\
        \midrule
        \textit{MountainCarContinuous-v0} & $87.8\pm4.8$ & $35.5\pm48.9$ & $30.8\pm42.0$ \\
        \textit{Swimmer-v3} & $66.4\pm13.7$ & $98.9\pm23.2$ & $92.8\pm28.2$\\ 
        \textit{Ant-v3} & $2,687.4\pm747.6$ & $2,558.8\pm1,140.0$ & $2,843.2\pm765.1$ \\
        \textit{FetchPickAndPlace-v1} & $0.948\pm0.022$ & $0.822\pm0.103$ & $0.853\pm0.150$\\
        \textit{walker-stand} & $891.3\pm143.2$ & $943.8\pm23.3$ & $978.6\pm2.2$\\
        \textit{finger-spin} & $928.4\pm31.7$ & $829.6\pm54.2$ & $925.5\pm25.9$\\ 
        \bottomrule
      \end{tabular}
  }
  \vspace{2em}
\end{table*}
In Table~\ref{tab:all_large}, we compare the performances of our methodology, $\pi_{L-only}$ (trained with SAC), and the symmetric HRL models implemented with two $\pi_{\omega_{large}}$ (i.e., no $\pi_{\omega_{small}}$ is used).  The objective of this analysis is to examine if the benefits of our method come from the direct use of HRL.  In other words, this analysis inspects if HRL offers unfair advantages to our methodology over $\pi_{L-only}$ trained with SAC.  For all tasks except \textit{Swimmer-v3}, it is observed that HRL offers little performance gain over the typical SAC method (i.e., $\pi_{L-only}$).  In some cases, HRL even exhibits performance drops, which are probably due to the instability during the training phase of HRL. For instance, in \textit{MountainCarContinuous-v0}, the model with two $\pi_{\omega_{large}}$ learn to complete the task in only two out of the five training runs, causing the relatively low average performance. On the other hand, Table~\ref{tab:all_large} also reveals that the agent benefits from the use of HRL in the case of \textit{Swimmer-v3}, such that it is able to achieve higher scores than the agents based on $\pi_{L-only}$.  Nevertheless, when compared with HRL with both $\pi_{\omega_{large}}$, our proposed asymmetric architecture with both $\pi_{\omega_{small}}$ and $\pi_{\omega_{large}}$ is able to reduce the costs further while maintaining the performance in \textit{Swimmer-v3}.

% models benefit from temporal abstraction with the use of HRL 

% and thus leading to inequitable comparison with SAC baselines. 

% For all tasks except \textit{Swimmer-v3}, there is little improvement of performance over the typical SAC method and some even exhibit performance drops, which may probably due to the unstable training of HRL. 

% For instance, in \textit{MountainCarContinuous-v0}, the models learn to complete the task only two out of five training runs, causing the relatively low average performance. These results also shows that temporal abstraction contributes to the performance of \textit{Swimmer-v3} and explicate its gaining of higher score. However, our model with both $\pi_{\omega_{small}}$ and $\pi_{\omega_{large}}$ is able to further reduce the costs while maintaining the performance in \textit{Swimmer-v3}.

\section{Analyses of the Baselines with More Data Samples}

In Table~\ref{tab:baseline_more_samples}, we show the training results of GAIL and BC with different numbers of data samples generated by the expert policy (i.e., $\pi_{L-only}$). The training data consist of different numbers of trajectories, where each trajectory contains all the state-action pairs collected in an episode. The network architecture for training these baselines is provided in Section~\ref{network_structure}. With more data samples from the expert, GAIL and BC gain improvement in their performances. Nevertheless, our model still outperform these baselines in four out of six tasks under the same level of computational costs.

% In Table~\ref{tab:baseline_more_samples}, we show the results of training using GAIL~\citep{gail} and BC~\citep{model_compression} with more data samples generated by expert policy $\pi_{L-only}$. The training data consist of different number of trajectories, and each trajectory contains all state-action pairs collected in an episode. The network architecure for training these baselines is illustrated in Section~\ref{network_structure}. With more data samples from the expert, GAIL and BC gain improvement in their performances. However, our model still outperform these baselines in four out of six tasks under the same level of computational costs.

\begin{table*}[!tb]
  \caption{Analyses of the baselines trained with more data samples from the expert model $\pi_{L-only}$.}
  \label{tab:baseline_more_samples}
  \centering
  \footnotesize
  \resizebox{\textwidth}{!}{
      \begin{tabular}{ *{7}{c} }
        \toprule
        & \multicolumn{2}{c}{\textbf{25 expert trajectories}} 
        & \multicolumn{2}{c}{\textbf{50 expert trajectories}} 
        & \multicolumn{2}{c}{\textbf{200 expert trajectories}} \\
        \cmidrule(lr){2-3}\cmidrule(lr){4-5}\cmidrule(lr){6-7}
        \raisebox{\dimexpr1.25\normalbaselineskip-.5\height}[0pt][0pt]{\begin{tabular}{@{}c@{}}
          Environment
        \end{tabular}} & GAIL & BC  & GAIL & BC & GAIL & BC \\
        \midrule
        \textit{MountainCarContinuous-v0} & $-99.9\pm0.0$ & $93.3\pm0.2$ & $-99.9\pm0.0$ & $93.2\pm0.2$ & $-99.9\pm0.0$ & $93.9\pm0.3$ \\
        \textit{Swimmer-v3} & $60.2\pm18.7$ & $72.8\pm3.7$ & $46.8\pm47.7$ & $81.4\pm1.1$ & $72.1\pm14.0$ & $76.1\pm6.5$\\
        \textit{Ant-v3} & $-893.0\pm1,046.5$ & $3,508.2\pm306.7$ & $81.1\pm249.1$ & $4,857.0\pm95.0$ & $765.2\pm636.2$ & $4,855.1\pm113.6$\\
        \textit{FetchPickAndPlace-v1} & $0.055\pm0.047$ & $0.054\pm0.023$ & $0.040\pm0.032$ & $0.056\pm0.015$ & $0.051\pm0.019$ & $0.294\pm0.164$\\
        \textit{walker-stand} & $651.4\pm189.1$ & $201.6\pm90.4$ & $680.8\pm167.9$ & $279.0\pm49.0$ & $676.5\pm61.1$ & $367.3\pm83.6$\\
        \textit{finger-spin} & $178.8\pm95.5$ & $168.5\pm162.1$ & $200.2\pm38.3$ & $489.0\pm158.6$ & $330.5\pm160.7$ & $516.7\pm343.1$\\ 
        \bottomrule
      \end{tabular}
  }
  \vspace{2em}
\end{table*}

\section{Computing Infrastructure}
In this section, we provide the configuration of our computing infrastructure in Table.~\ref{tab:infrastructure} for reference.

\section{Reproducibility}
We implemented the proposed methodology based on the codes from Stable Baselines~\citep{stable-baselines} and RL Baselines Zoo~\citep{rl-zoo}, which are high-quality implementations for RL algorithms and training models. We modified the source codes for DQN, SAC, and the training procedure to adapt to our methodology. The source codes for our experiments are well verified, and all the experiments in our paper are fully reproducible. Please refer to the following github repository for more detailed instructions: \textcolor{blue}{\href{https://github.com/anonymouscjc/Computational-Cost-Aware-Control-Using-Hierarchical-Reinforcement-Learning}{link}}.
% We implemented the proposed MB-DQN based on the RLTF framework. RLTF is a research framework that provides high-quality implementations of common RL algorithms based on TensorFlow deveplopment platform. We modified the source code of RLTF and added an additional option \textit{MB-DQN} in DQN family. All the conducted experiments presented in our paper are re-producible with easy-following instructions. For more details about the provided source coded, please refer to the anonymous github repository as following link: \href{https://github.com/NIPS-Paper-139/MB-DQN}{https://github.com/NIPS-Paper-139/MB-DQN}.
\begin{table}[!tb]
\centering
\renewcommand{\arraystretch}{1.3}
\caption{Specification of our computing infrastructure.}
\small
\resizebox{\linewidth}{!}{
\begin{tabular}{l|l}

\toprule
Component           & Customized Machine \\
\midrule
Processor        & 32 cores / 64 threads (3.0GHz, up to 4.2GHz) \\
Hard Disk Drive  & 6TB SATA3 7200rpm \\
Solid-State Disk & 1TB PCIe Gen 3 NVMe \\
Graphics Card    & NVIDIA GeForce$^{\circledR}$ RTX 2080Ti (two per instance) \\
Memory           & 16GB DDR4 2400MHz (128GB in total) \\
\bottomrule

\end{tabular}
}
\label{tab:infrastructure}

\end{table}
% \clearpage

\bibliographystyle{unsrtnat}
% \bibliography{neurips_2020}
\bibliography{reference}